\documentclass[12pt,a4paper]{article}
\usepackage[utf8]{inputenc}

\usepackage{color} 

\definecolor{human.100}{RGB}{165, 30, 55}
\definecolor{vgg.100}{RGB}{0, 105, 170}
\definecolor{googlenet.100}{RGB}{80, 170, 200}
\definecolor{alexnet.100}{RGB}{65, 90, 140}

\usepackage{amsmath}
\usepackage{amsfonts}
\usepackage{amssymb}
\usepackage{graphicx}
\usepackage{subfig}
\usepackage[colorlinks=true,citecolor=black,linkcolor=black,urlcolor=black]{hyperref}
\usepackage{apacite}
\usepackage{authblk}
\usepackage{float}
\usepackage{todonotes}
\usepackage{fullpage}
\usepackage[font={small,it}]{caption}
\usepackage{lineno}
\usepackage{booktabs}
\usepackage{eurosym}
\usepackage[margin=10pt,font=small,labelfont=bf]{caption}
\usepackage{multirow,booktabs,setspace,caption}
\usepackage{scalerel,stackengine}
\usepackage{rotating}
\DeclareCaptionLabelSeparator*{spaced}{\\[2ex]}
\graphicspath{{figures/}}

\author[1,2]{Robert Geirhos}
\author[1,3]{David H. J. Janssen}
\author[1,4,5]{Heiko H. Sch\"{u}tt}
\author[2,3]{Jonas~Rauber}
\author[2,6,7]{Matthias Bethge}
\author[1,2,6,8]{Felix A. Wichmann}

\affil[1]{Neural Information Processing Group, University of T\"{u}bingen, Germany}
\affil[2]{Centre for Integrative Neuroscience, University of T\"{u}bingen, Germany}
\affil[3]{Graduate School of Neural Information Processing, University of T\"{u}bingen, Germany}
\affil[4]{Graduate School of Neural and Behavioural Sciences, University of T\"{u}bingen, Germany}
\affil[5]{Department of Psychology, University of Potsdam, Germany}
\affil[6]{Bernstein Center for Computational Neuroscience, T\"{u}bingen, Germany}
\affil[7]{Max Planck Institute for Biological Cybernetics, T\"{u}bingen, Germany}
\affil[8]{Max Planck Institute for Intelligent Systems, T\"{u}bingen, Germany}

\title{
Comparing deep neural networks against humans: object recognition when the signal gets weaker\footnote{Please note that this paper has been greatly extended to an article published at NeurIPS 2018. Unless referring to specific results contained here, we recommend reading and citing the new paper (Geirhos et al., 2018).}
}

\begin{document}

\date{} %
\maketitle

\abstract{
Human visual object recognition is typically rapid and seemingly effortless, as well as largely independent of viewpoint and object orientation. Until very recently, animate visual systems were the only ones capable of this remarkable computational feat. This has changed with the rise of a class of computer vision algorithms called deep neural networks (DNNs) that achieve human-level classification performance on object recognition tasks. Furthermore, a growing number of studies report similarities in the way DNNs and the human visual system process objects, suggesting that current DNNs may be good models of human visual object recognition. Yet there clearly exist important architectural and processing differences between state-of-the-art DNNs and the primate visual system. The potential behavioural consequences of these differences are not well understood. We aim to address this issue by comparing human and DNN generalisation abilities towards image degradations.
We find the human visual system to be more robust to image manipulations like contrast reduction, additive noise or novel eidolon-distortions. In addition, we find progressively diverging classification error-patterns between humans and DNNs when the signal gets weaker, indicating that there may still be marked differences in the way humans and current DNNs perform visual object recognition. We envision that our findings as well as our carefully measured and freely available behavioural datasets\footnote{Data and materials available at \url{https://github.com/rgeirhos/object-recognition}} provide a new useful benchmark for the computer vision community to improve the robustness of DNNs and a motivation for neuroscientists to search for mechanisms in the brain that could facilitate this robustness.}

\newpage
\nocite{Geirhos2018generalisation}

\section{Introduction}
  \label{Introduction}

The visual recognition of objects by humans in everyday life is typically rapid and effortless, as well as largely independent of viewpoint and object orientation \cite<e.g.>{Biederman_1987}. This ability of the primate visual system has been termed \textit{core object recognition}, and much research has been devoted to understanding this process \cite<see>[for a review]{DiCarlo2012}. We know, for example, that it is possible to reliably identify objects in the central visual field within a single fixation in less than 200 ms when viewing ``standard'' images \cite{DiCarlo2012, Potter1976, Thorpe1996}. Based on the rapidness of object recognition, core object recognition is often hypothesized to be achieved with mainly feedforward processing although feedback connections are ubiquitous in the primate brain \cite<but see, e.g.>[for a critical assessment of this argument]{Gerstner_2005}. Object recognition is believed to be realized by the ventral visual pathway, a hierarchical structure consisting of the areas V1-V2-V4-IT, with information from the retina reaching the cortex in V1 \cite<e.g.>{goodale1992}. Although aspects of this process are known, others remain unclear.
Until very recently, animate visual systems were the only known systems capable of visual object recognition. This has changed, however, with the advent of brain-inspired deep neural networks (DNNs) which, after having been trained on millions of labeled images, achieve human-level performance when classifying objects in images of natural scenes \cite{Krizhevsky2012}. DNNs are now employed on a variety of tasks and set the new state-of-the-art, sometimes even surpassing human performance on tasks which were a few years ago thought to be beyond an algorithmic solution for decades to come \cite{He2015delving,silver2016}. For an excellent introduction to DNNs see e.g. \citeA{LeCun2015}.

Although being in the first place an engineering discipline, the field of computer vision (interested in designing algorithms and building machines that can see) has always been interested in human vision: As in object recognition, our visual system is often remarkably successful, acting as de facto performance benchmark for many tasks. It is thus not surprising that there has always been an exchange between researchers in computer vision and human vision, such as the design of low-level image representations \cite{Simoncelli_etal_1992,Simoncelli_Freeman_1995} and the investigation of underlying coding principles such as redundancy reduction \cite{Atick1992, Barlow1961, Olshausen1996}. With the advent of DNNs over the course of the last few years, this exchange has even deepened. It is thus not surprising that some studies have started investigating similarities between DNNs and human vision, drawing parallels between network and biological units or network layers and visual areas in the primate brain. Clearly, describing network units as biological neurons is an enormous simplification given the sophisticated nature and diversity of neurons in the brain \cite{Douglas_Martin_1991}. Still, often the strength of a model lies not in replicating the original system but rather in its ability to capture the important aspects while abstracting from details of the implementation \cite<e.g.>{Kriegeskorte2015}.

\subsection{Behavioural comparison between humans and DNNs}
Thorough comparisons of human and DNN \text{behaviour} have been relatively rare. Behaviour goes well beyond overall performance: It comprises all performance changes as a function of certain stimulus properties, e.g. how classification accuracy depends on image background and contrast or the type and distribution of errors. Ideally, computational models of behaviour should not only be able to predict the overall accuracy of humans, but be able to describe behaviour on a more fine-grained level, e.g. in the current experiment on a category-by-category level. The ultimate goal should be the prediction of behaviour on a trial-by-trial basis, termed \emph{molecular psychophysics} \cite{Green_1964,Schonfelder_Wichmann_2012}.
An important early step into comparing human and DNN behaviour was the work of \citeA{Lake2015} reporting that DNNs are able to predict human category typicality ratings for images. Another study by \citeA{Kheradpisheh2016} found largely similar performance on view-invariant, background-controlled object recognition and, for some DNNs, highly similar error distributions.
On the other hand, so-called \textit{adversarial examples} have cast some doubt on the idea of broad-ranging manlike DNN behaviour. For any given image it is possible to perturb it minimally in a principled way such that DNNs mis-classify it as belonging to an arbitrary other category \cite{Szegedy2014}. This slightly modified image is then called an adversarial example, and the manipulation is imperceptible to human observers \cite{Szegedy2014}. 

The ease at which DNNs can be fooled speaks to the need of a careful, psychophysical comparison of human and DNN behaviour. As the possibility to systematically search for adversarial examples is very limited in humans, it is not known how to quantitatively compare the robustness of humans and machines against adversarial attacks. However, other behavioural measurements are known to have contributed much to our current understanding of the human visual system:
Psychophysical investigations of human behaviour on object recognition tasks, measuring accuracies depending on image colour (grayscale vs. colour), image contrast  and the amount of additive visual noise have been powerful means of exploring the human visual system, revealing much about the internal computations and mechanisms at work \cite<e.g.>{Nachmias_Sansbury_1974,Pelli_Farell_1999,Wichmann_1999,Henning_etal_2002b,Carandini2012, Carandini1997, Delorme2000}. As a consequence, similar experiments might yield equally interesting insights into the functioning of DNNs, especially as a comparison to human behaviour. In this study, we obtain and analyse human and DNN classification data for the three above-mentioned, well-known image degradations. In addition, we employ a novel image manipulation method. The stimuli generated by the so-called eidolon-factory \cite{Koenderink2017} are parametrically controlled distortion of an image. Eidolons aim to evoke similar visual awareness as objects perceived in the periphery, giving them some biological justification. To our knowledge, we are among the first to measure DNN performance on these tasks and compare their behaviour to carefully measured human data, in particular using a controlled lab environment (instead of Amazon Mechanical Turk without sufficient control about presentation times, display calibration, viewing angles, and sustained attention of participants). 

In this study, we employ a paradigm\footnote{This is the same paradigm as reported by \citeA{Wichmann2017}.} aimed at comparing human observers and DNNs as fair as possible using an image categorization task with short presentation times (200 ms) along with backwards masking by a high-contrast 1/\textit{f} noise mask, known to minimize, as much as psychophysically possible, feedback influence in the brain. This is important since all investigated networks rely on purely feedforward computations. We perform psychophysical experiments on both human observers and DNNs to assess how robust the three currently well-known DNNs AlexNet \cite{Krizhevsky2012}, GoogLeNet \cite{Szegedy2015} and VGG-16 \cite{Simonyan2015} are towards image degradations in comparison to human participants. %

DNNs provide exciting new opportunities for computational modelling of vision---and we envisage DNNs to have a major impact on our understanding of human vision in the future, essentially agreeing with assessments voiced by \citeA{Kriegeskorte2015}, \citeA{Kietzmann_etal_2017} and \citeA{VanRullen_2017}. With this study, we aim to shed light on the behavioural consequences of the \emph{currently existing} architectural, processing and training differences between the tested DNNs and the primate brain. We envision that our analyses as well as our carefully measured and freely available behavioural datasets (\url{https://github.com/rgeirhos/object-recognition}) may provide a new useful benchmark for the computer vision community to improve the robustness of DNNs and a motivation for neuroscientists to search for mechanisms in the brain that could facilitate human robustness.

\section{Methods}
  \label{methods}
\subsection{General}
We tested four ways of degrading images: conversion to grayscale, reducing image contrast, adding uniform white noise, and increasing the strength of a novel image distortion from the eidolon toolbox \cite{Koenderink2017}.
Here we give an overview about the experimental procedure and about the observers and deep neural networks that performed these experiments. In the Appendix we provide details on the categories and image database used (Section~\ref{methods:categories}), as well as information about image preprocessing (Section~\ref{methods: image_preprocessing}), including plots of example stimuli at different levels of signal strength. In Section~\ref{methods: apparatus} of the Appendix we list the specifics of our experimental setup; for now it might be enough to know that images in the psychophysical experiments were always displayed at the center of the screen at a size of $3 \times 3$ degrees of visual angle.

\subsection{Procedure}
\begin{figure}[t]
\centering
 \includegraphics[width=0.9\textwidth]{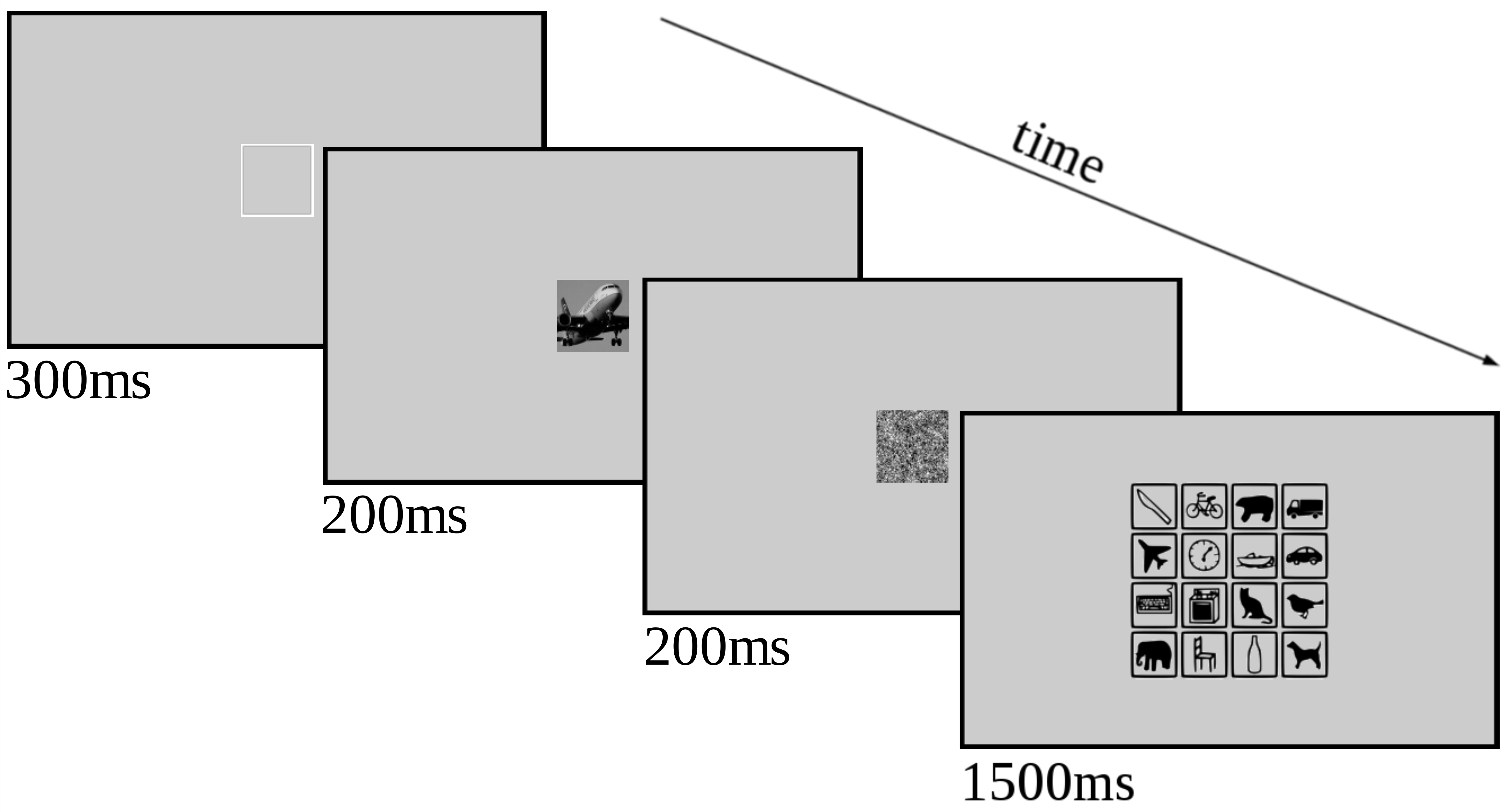}
\caption{Schematic of a trial. After the presentation of a central fixation square (300 ms), the image was visible for 200 ms, followed immediately by a noise-mask with 1/\textit{f} spectrum (200 ms). Then, a response screen appeared for 1500 ms, during which the observer clicked on a category. Note that we increased the contrast of the noise-mask in this figure for better visibility when printed.
Categories row-wise from top to bottom: \texttt{knife, bicycle, bear, truck, airplane, clock, boat, car, keyboard, oven, cat, bird, elephant, chair, bottle, dog}. The icons are a modified version of the ones from the MS COCO website (\url{http://mscoco.org/explore/}).}
\label{fig:typical_trial}
\end{figure}

In each trial a fixation square was shown for 300 ms, followed by an image shown for only 200 ms, in turn immediately followed by a full-contrast pink noise mask (1/\textit{f} spectral shape) of the same size and duration. Participants had to choose one of 16 entry-level categories (see Section~\ref{methods:categories} for details on these categories) by clicking on a response screen shown for 1500 ms\footnote{During practice trials the response screen was visible for another 300 ms in case an incorrect category was selected, and along with a short low beep sound the correct category was highlighted by setting its background to white.}. During the whole experiment, the screen background was set to a grey value of 0.454 in the [0, 1] range, corresponding to the mean grayscale value of all images in the dataset (41.17 cd/m\textsuperscript{2}). Figure~\ref{fig:typical_trial} shows a schematic of a typical trial.

Prior to starting the experiment, all participants were shown the response screen and asked to name all categories to ensure that the task was fully clear. They were instructed to click on the category that they thought resembles the image best, and to guess if they were unsure. They were allowed to change their choice within the 1500 ms response interval; the last click on a category icon of the response screen was counted as the answer. The experiment was not self-paced, i.e. the response screen was always visible for 1500 ms and thus, each experimental trial lasted exactly 2200 ms (300 ms + 200 ms + 200 ms + 1500 ms).

On separate days we conducted four different experiments with 1,280 trials per participant each (eidolon-experiment: three sessions of 1,280 trials each). In the colour-experiment, we used two distinct conditions (colour vs. grayscale), whereas in the contrast-experiment and in the noise-experiment eight conditions were explored (corresponding to eight different contrast values or noise power densities, respectively). In the eidolon-experiment, 24 distinct conditions were employed. For each experiment, we randomly chose 80 images per category from the pool of images without replacement (i.e., no observer ever saw an image more than once throughout the entire experiment). Within each category, all conditions were counterbalanced. Stimulus selection was done individually for each participant to reduce the probability of an accidental bias in the image selection. Images within the experiments were presented in randomized order. After 256 trials (colour-experiment, noise-experiment and eidolon-experiment) and 128 trials (contrast-experiment), the mean performance of the last block was displayed on the screen, and observers were free to take a short break. The total time necessary to complete all trials was 47 minutes per session, not including breaks and practice trials. In total, the results reported in this article are based on 39,680 psychophysical trials. Ahead of each experiment, all observers conducted approximately 10 minutes of practice trials to gain familiarity with the task and the position of the categories on the response screen.

\subsection{Observers and deep neural networks}
Three observers participated in the colour-experiment (all male; 22 to 28 years; mean: 25 years). In each of the other experiments, five observers took part (contrast-experiment and noise-experiment: one female, four male; 20 to 28 years; mean: 23 years. Eidolon-experiment: three female, two male; 19 to 28 years; mean: 22 years). Subject-01 is an author and participated in all but the eidolon-experiment. All other participants were either paid \EUR{10} per hour for their participation or gained course credit. All observers were students of the University of T\"{u}bingen and reported normal or corrected-to-normal vision.

We used three DNNs for our analysis: AlexNet \cite{Krizhevsky2012}, GoogLeNet \cite{Szegedy2015} and VGG-16 \cite{Simonyan2015}. All three networks were specified within the Caffe framework \cite{Jia2014} and acquired as a pre-trained model. VGG-16 was obtained from the Visual Geometry Group's website (\url{http://www.robots.ox.ac.uk/~vgg/}); AlexNet and GoogLeNet from the BLVC model zoo website (\url{https://github.com/BVLC/caffe/wiki/Model-Zoo}). We reproduced the respective specified accuracies on the ILSVRC 2012 validation dataset in our setting.

All DNNs require images to be specified using RGB planes; to evaluate the performance using grayscale images we stacked a grayscale image three times in order to obtain the desired form specified by the caffe.io module (\url{https://github.com/BVLC/caffe/blob/master/python/caffe/io.py}). Images were fed through the networks using a single feedforward pass of the $224 \times 224 $ pixels center crop.

\section{Results}
  \label{results}
  
\newcommand\xWidth{0.4}

Trials in which human observers failed to click on any category were recorded as an incorrect answer in the data analysis, and are shown as a separate category (top row) in the confusion matrices (DNNs, obviously, never fail to respond). Such a failure to respond occurred in only 1.2\% of all trials, and did not differ meaningfully between the different experiments.
The terms 'accuracy' and 'performance' are used interchangeably. All data, if not stated otherwise, were analyzed using R version 3.2.3 \cite{RCoreteam}.

\subsection{Accuracy and response distribution entropy}
\label{results_general}

When showing \emph{accuracy} in any of the plots, the error bars provided have two distinct meanings: First, for DNNs they indicate the range of DNN accuracies resulting from seven\footnote{Seven runs are the maximum possible number of runs without ever showing an image to a DNN more than once per experiment.} runs on different images, with each run consisting of the same number of images per category and condition that a single human observer was exposed to. This serves as an estimate of the variability of DNN accuracies as a function of the random choice of images. Second, the error bars for human participants likewise correspond to the range of their accuracies (not the often shown S.E. of the means, which would be much smaller).

In addition we assessed the \emph{response distribution entropy} of humans and DNNs as a function of image degradation strength. Entropy is a measure quantifying how close a distribution is to the uniform distribution (the higher the entropy, the closer it is). The distribution obtained by throwing a fair die many times should therefore have higher entropy than the distribution obtained from repeatedly throwing a rigged die. In the context of our experiments, it is used to measure whether observers or DNNs exhibit a bias towards certain categories: if so, the response distribution entropy will be lower than the maximal value of 4 bits (given 16 categories). We calculated the Shannon entropy H of response distribution $\mathcal{X}$ as follows:\\
	H$(\mathcal{X}) = - \sum_{i=1}^{16}p(x_i) log_2(p(x_i))$, with $p(x_i)$ being the fraction of responses for category $i$ (e.g. $p(x_{car})=0.25$ if an observer responds \texttt{car} every fourth trial on average).

\subsubsection{Colour-experiment}
\label{exp:colour}
\begin{figure}[ht]
	\centering
	\subfloat[][Colour-experiment accuracy]{\includegraphics[width=\xWidth\textwidth]{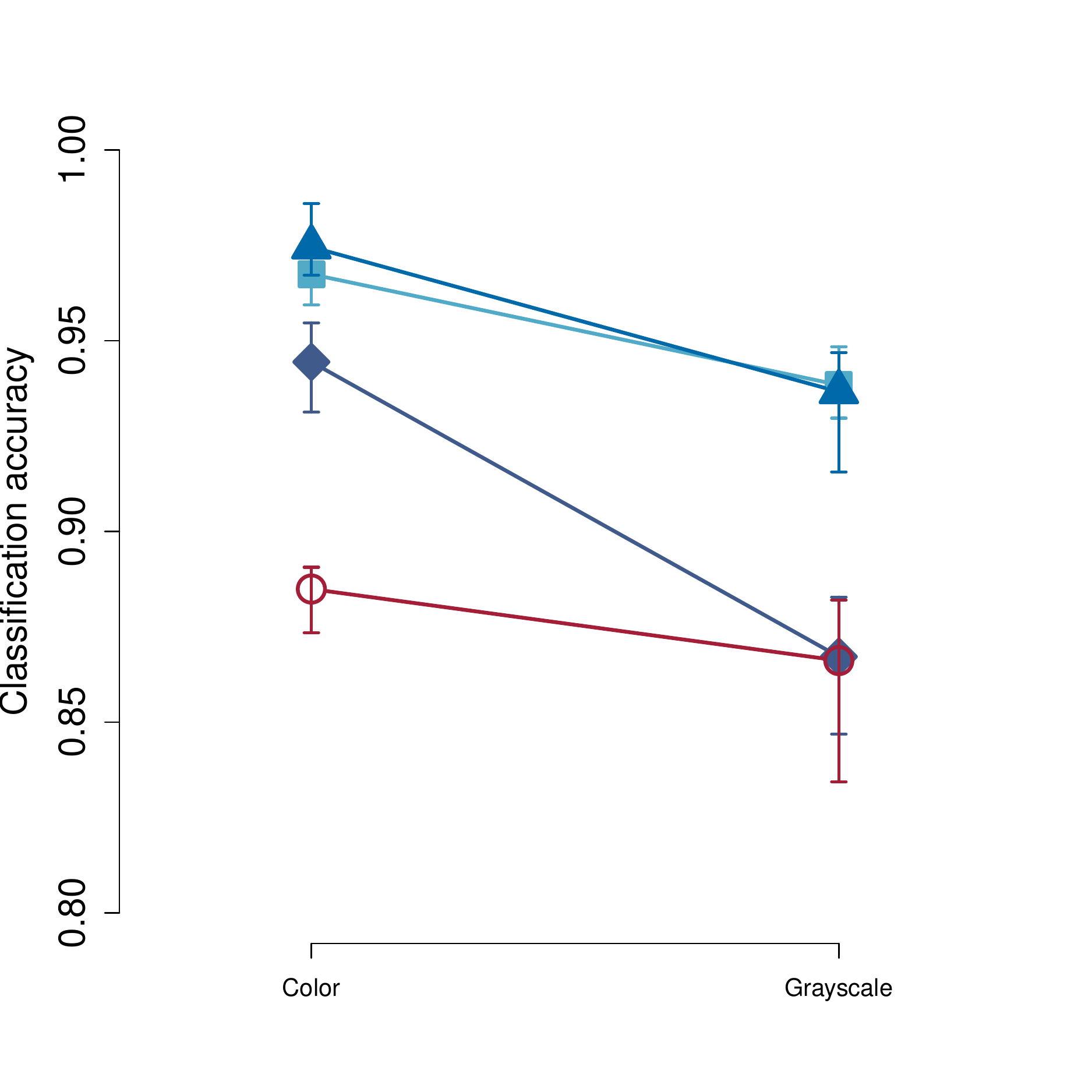}}
	\subfloat[][Colour-experiment entropy]{\includegraphics[width=\xWidth\textwidth]{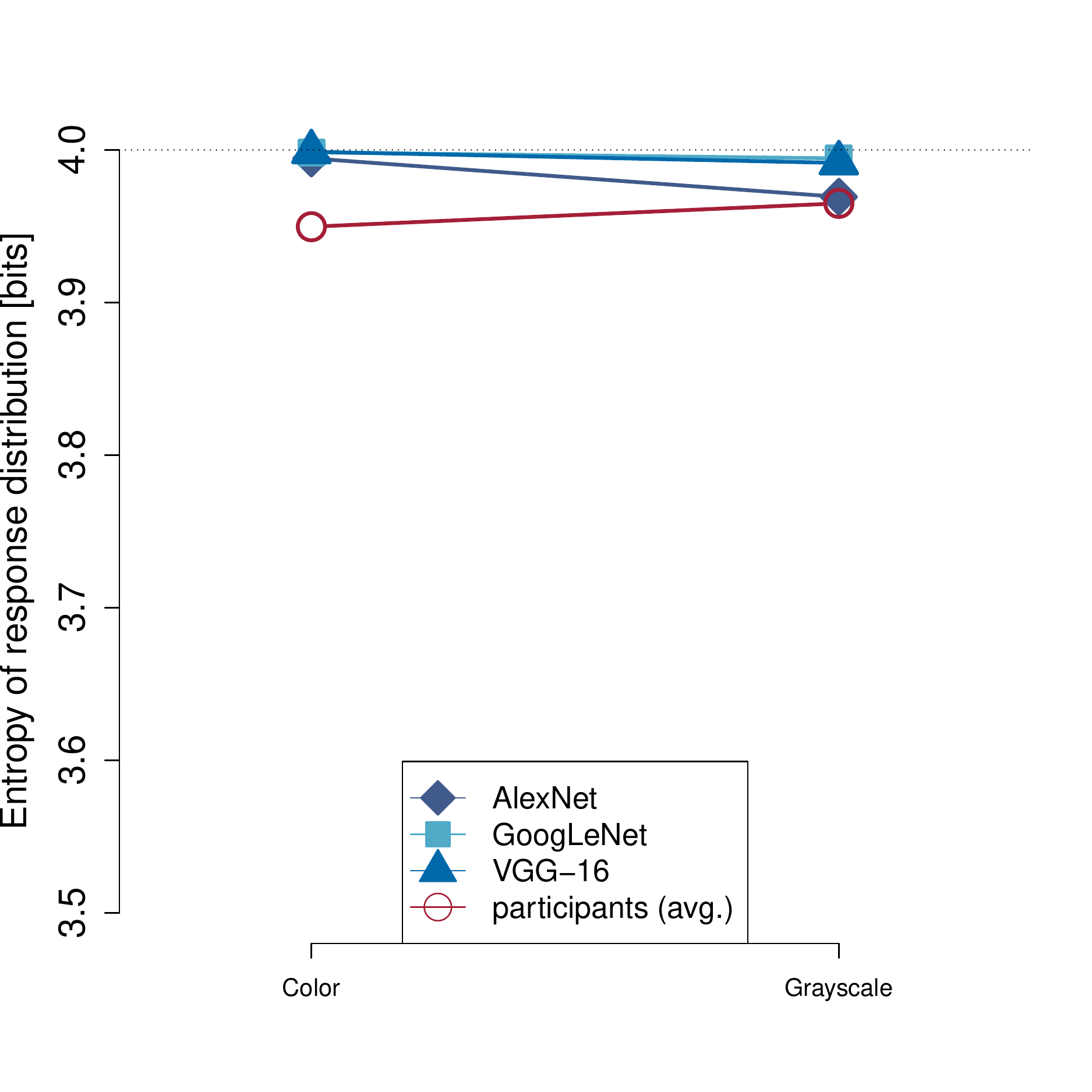}}
	\caption{Results for the colour-experiment (n=3). \textbf{(a)} \emph{Accuracy}. DNNs are shown in shades of blue, human data in red; diamonds correspond to \textcolor{alexnet.100}{AlexNet}, squares to \textcolor{googlenet.100}{GoogLeNet}, triangles to \textcolor{vgg.100}{VGG-16}, and circles to \textcolor{human.100}{human observers}; error bars as described in section~\ref{results_general}. \textbf{(b)} \emph{Response distribution entropy}. Plotting conventions as in (a).}
	\label{fig:colour_performance}
\end{figure}

We conducted a paired-samples \textit{t}-test to assess the difference in accuracy between coloured and grayscale images for each network and observer (Table~\ref{tab:colour_difference} in the Appendix). In order to account for multiple comparisons, the critical significance level of .05 was adjusted to $\frac{.05}{6} = .008\overline{3}$ by applying Bonferroni correction. As shown in Figure~\ref{fig:colour_performance}(a) all three networks performed significantly worse for grayscale images compared to coloured images (4.81\% drop in performance on average: significant, but not dramatic in terms of effect size). Human observers, on the other hand, did not show on average a significant reduction in accuracy (only $ 1.88\% $ accuracy drop for grayscale images). As can be seen from the range of human grayscale results, obsververs differed in their ability to cope with grayscale images.

The response distribution entropy shown in Figure~\ref{fig:colour_performance}(b) is innocuous: The DNNs distributed their responses perfectly among the 16 categories, and human observers are only marginally worse.

\subsubsection{Contrast-experiment}
\begin{figure}
    \centering
	\subfloat[][Contrast-experiment accuracy]{\includegraphics[width=\xWidth\textwidth]{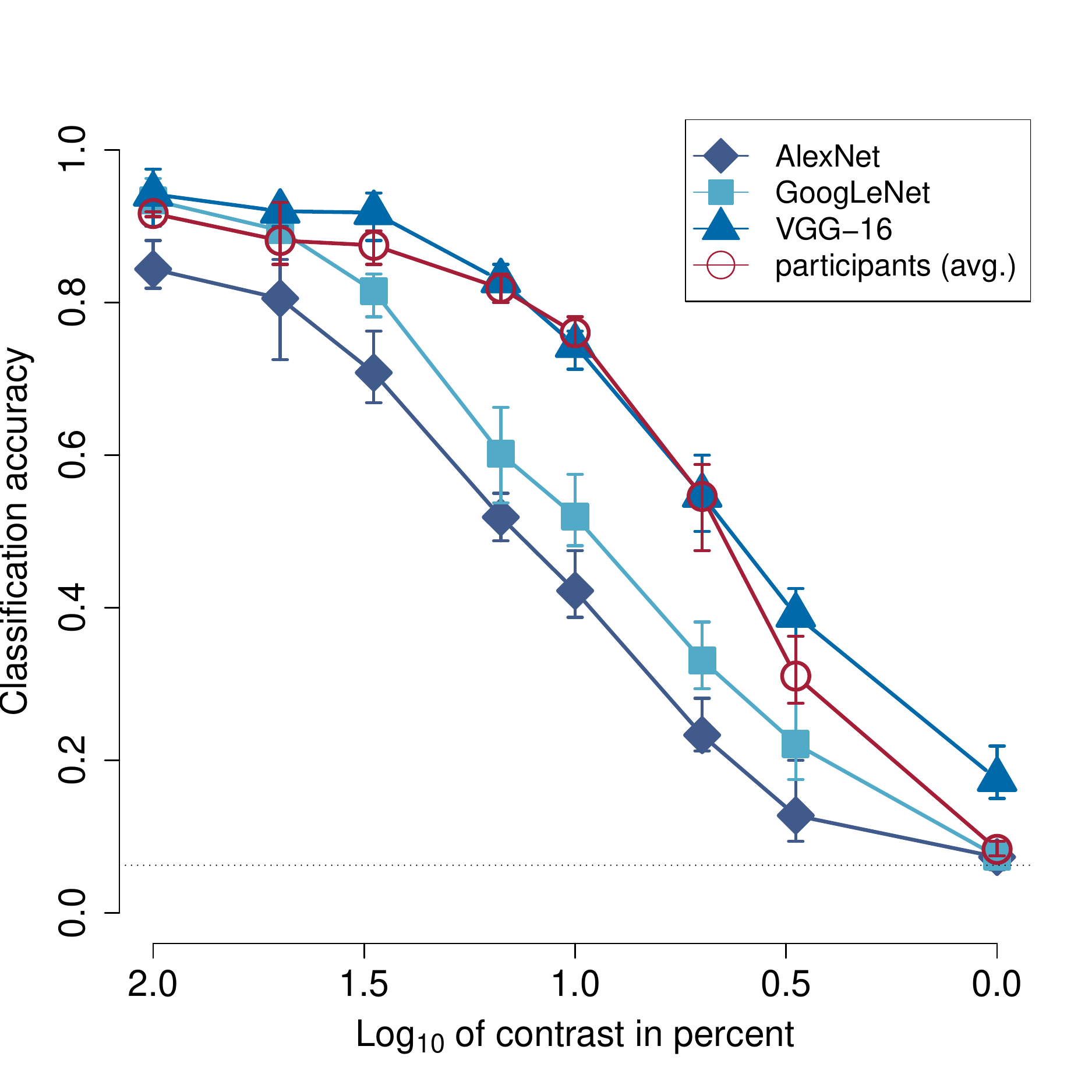}}
	\subfloat[][Contrast-experiment entropy]{\includegraphics[width=\xWidth\textwidth]{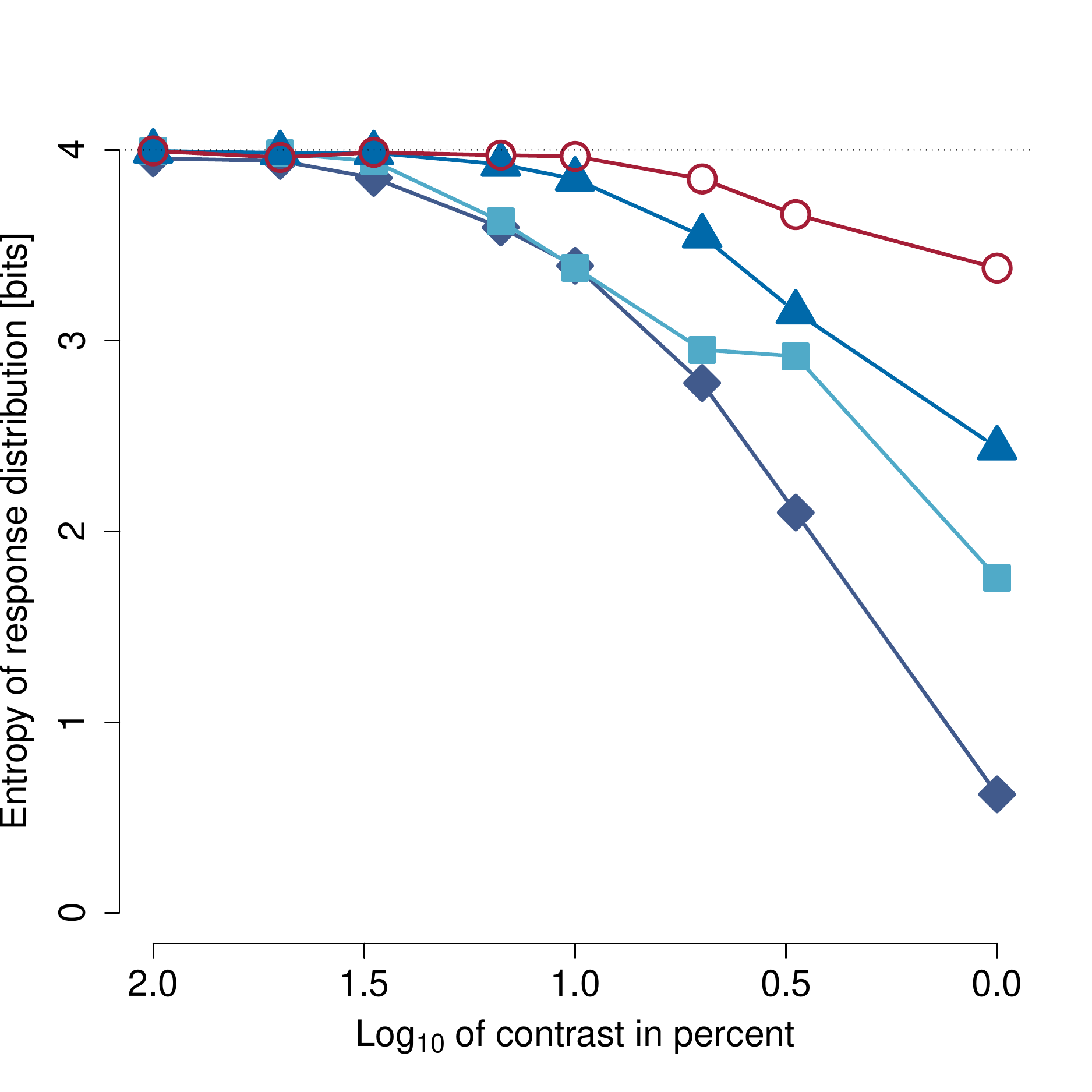}}
	\hfill
	\subfloat[][Noise-experiment accuracy]{\includegraphics[width=\xWidth\textwidth]{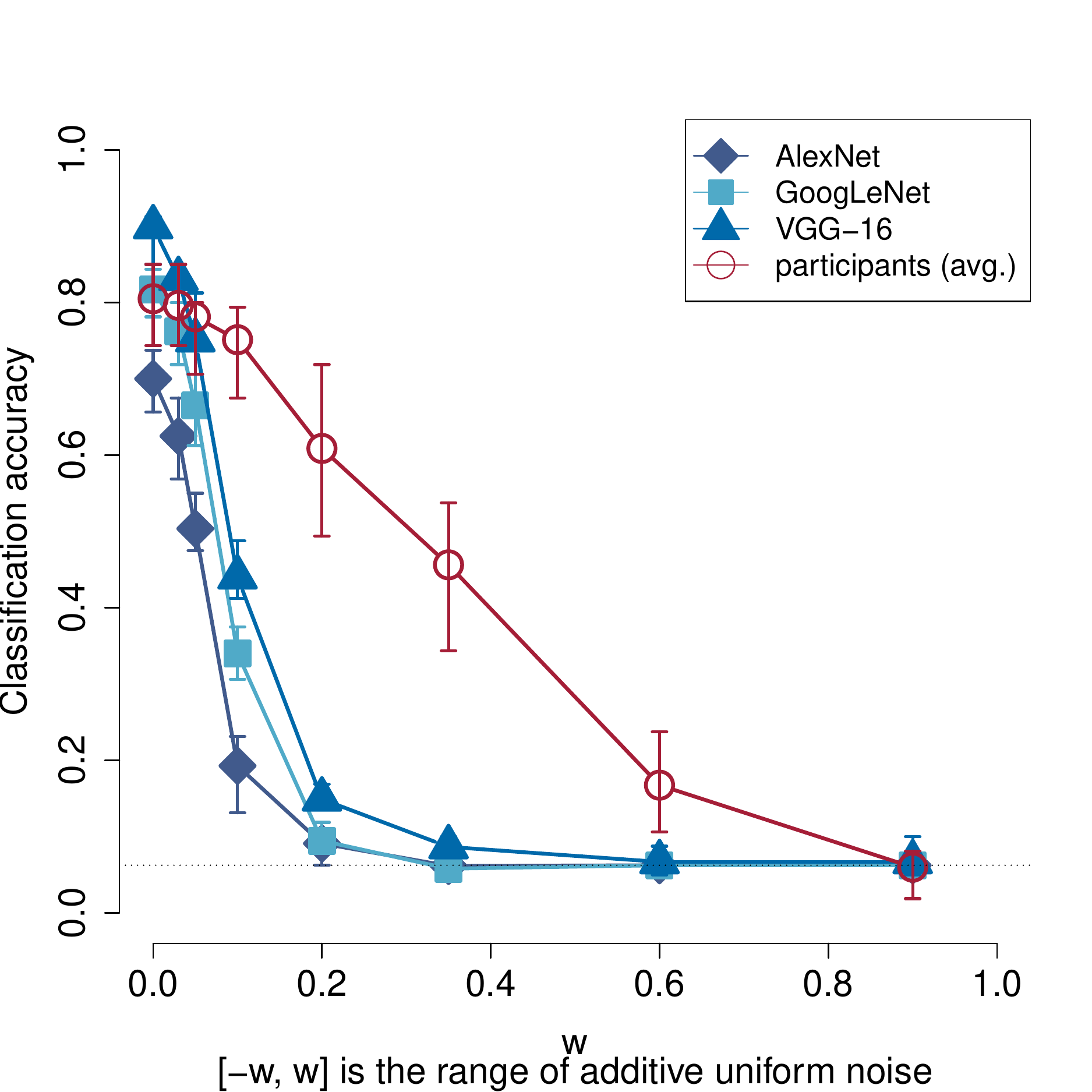}}
	\subfloat[][Noise-experiment entropy]{\includegraphics[width=\xWidth\textwidth]{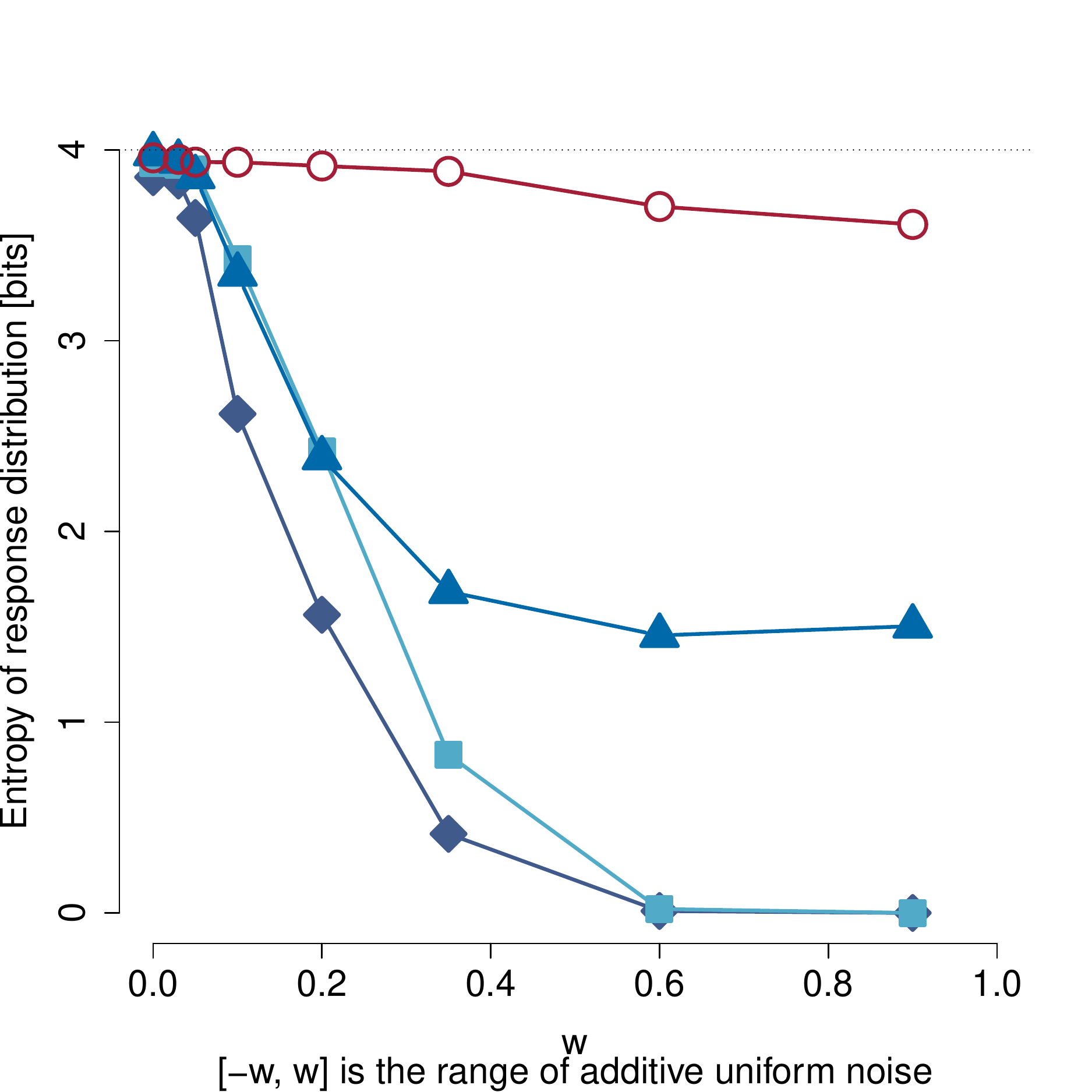}}
	\hfill
	\subfloat[][Eidolon-experiment accuracy (coherence parameter = 1.0)]{\includegraphics[width=\xWidth\textwidth]{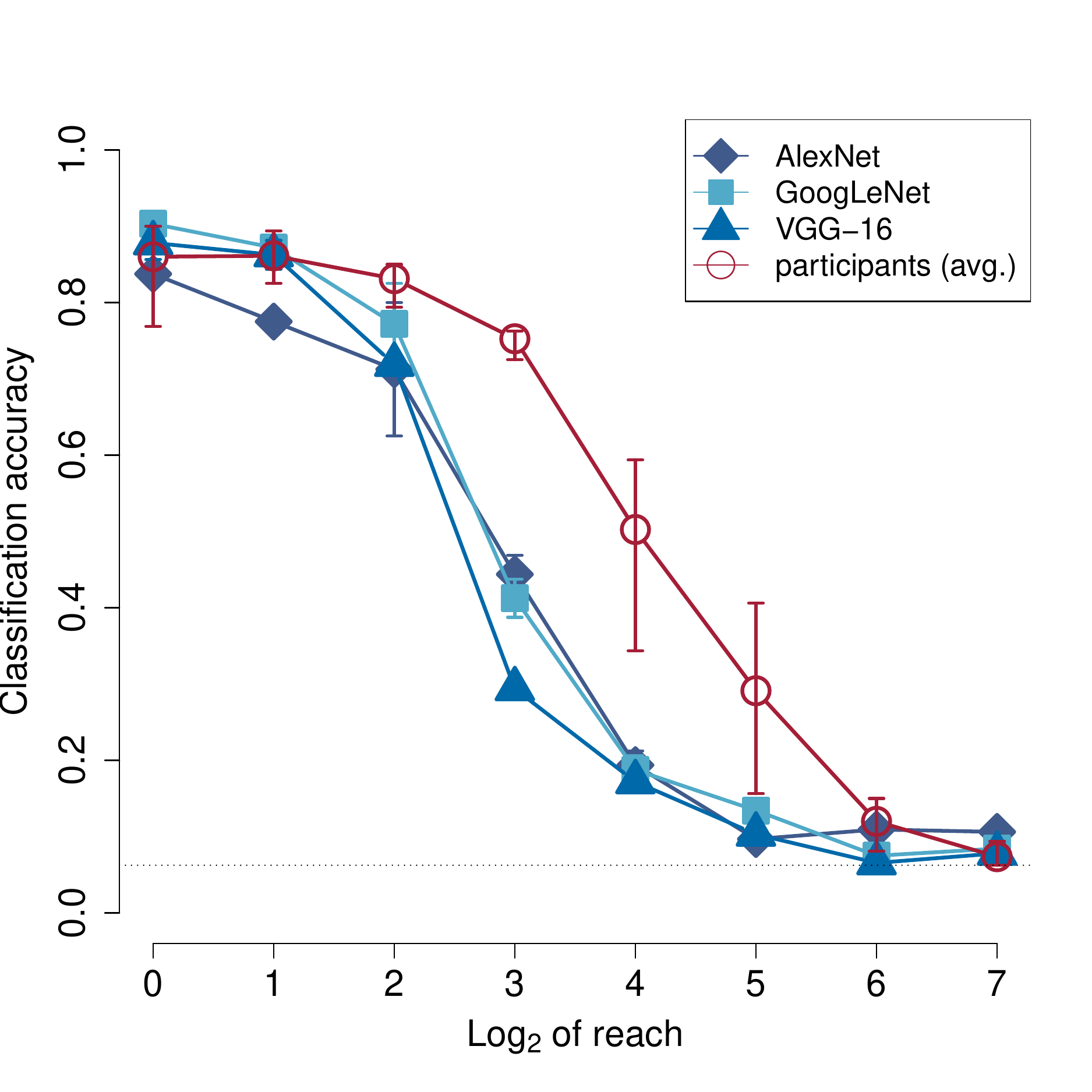}}
	\subfloat[][Eidolon-experiment entropy]{\includegraphics[width=\xWidth\textwidth]{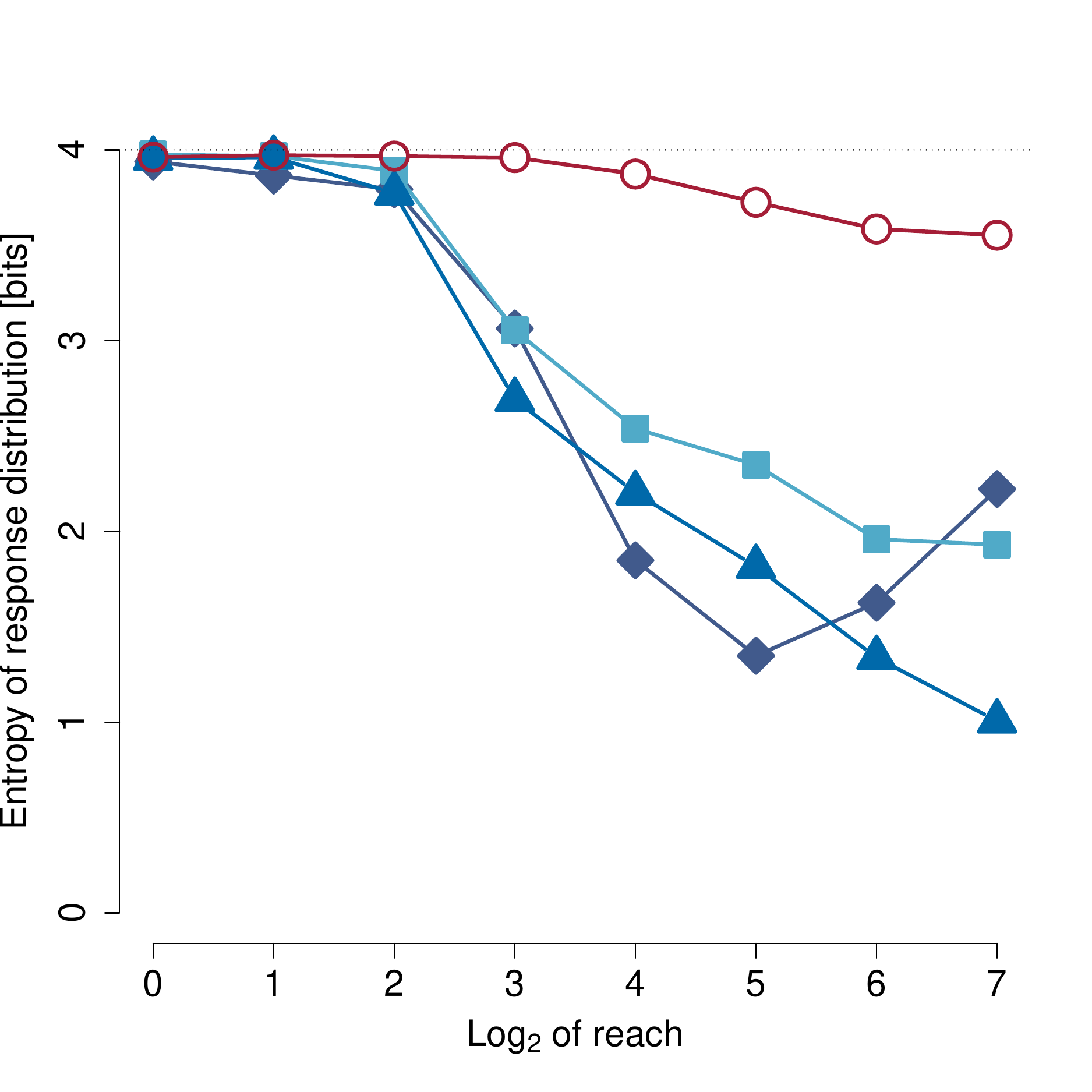}}
	\caption{Results for contrast-, noise- and eidolon-experiment (n=5 each).\\ \textbf{(a, c, e)}(Left) \emph{Accuracy}. Plotting conventions as in Figure~\ref{fig:colour_performance}.\\
	\textbf{(b, d, f)}(Right) \emph{Response distribution entropy}.}
	\label{fig:accuracy_entropy}
\end{figure}

As shown in Figure~\ref{fig:accuracy_entropy}(a), accuracies for the contrast-experiment ranged from approximately $ 91-94\%$ (VGG-16, GoogLeNet and human average) and $ 84\%$ (AlexNet) for full contrast to chance level ($ \frac{1}{16} = 6.25 \% $) for 1\% of contrast, except for VGG-16 which still achieves 17.5\% correct responses. AlexNet's and GoogLeNet's performance dropped more rapidly than human and VGG-16's performance for lower contrast levels.

The response distribution entropy shown in Figure~\ref{fig:accuracy_entropy}(b) reveals, however, that all three DNNs showed an increasing bias towards few categories (in other words, they did no longer distribute their responses evenly among the 16 categories if the contrast was lowered). Human observers, on the other hand, still largely distributed their responses sensibly across the 16 categories.

\subsubsection{Noise-experiment}
\label{exp:noise}
The data for the noise-experiment were analyzed in the same way as the contrast-experiment data. Overall, we found drastic differences in classification accuracy, with human observers clearly outperforming all three networks. As can be seen in Figure~\ref{fig:accuracy_entropy}(c), by increasing the noise width from $ 0.0 $ (no noise) to $ 0.1 $, VGG-16's performance drops from an accuracy of $ 89.91\% $ to $ 44.02\% $; GoogLeNet's drops from $ 81.70 \%$ to $ 34.02\% $ and AlexNet's from $ 70.00\% $ to $ 19.29\% $. Human observers, on the other hand, only drop from $ 80.50\% $ to $ 75.13\% $.

The response distribution entropy shown in Figure~\ref{fig:accuracy_entropy}(d) shows again that all of the investigated DNNs exhibit a strong bias towards few categories if the images contained additive noise. For AlexNet and GoogLeNet, the response distribution entropy is close to 0 bits for a noise width of 0.6 or more, which means that they responded with a single category for these images (category \texttt{bottle} for both). Interestingly, these preferred categories are usually not the same across experiments or networks (Figures~\ref{fig:noise_confusion} and \ref{fig:contrast_confusion}), and they do not simply match the probabilities of the categories in the ImageNet training database. The network responses therefore are not converging to their prior distribution, which would be a sensible way to behave in the absence of a signal. Human observers, as with low contrast, largely distributed their responses evenly across the 16 categories.

\subsubsection{Eidolon-experiment}

Results for the eidolon-experiment with maximal coherence of $1.0$ are shown in Figure~\ref{fig:accuracy_entropy}(e) and (f). The complete results of the eidolon-experiment for all coherence settings are provided in the Appendix, Figure~\ref{fig:eidolon_performance}. In terms of accuracy, network and human performance naturally were  approximately equal for very low values of reach (no distortion, therefore high accuracies) and for very high values of reach (heavy distortion, accuracy at chance level). In the range between these extremes, their accuracies followed the typically observed s-shaped pattern known from most psychophysical experiments varying a single parameter. However, human observers clearly achieved higher accuracies than all three networks for intermediate distortions. In the full coherence case, the largest difference between network and human performance was observed for a reach value of $ 2^3 = 8 $ (38.3\% network accuracy vs. 75.3\% human accuracy, averaged across networks and observers). The coherence-parameter, albeit having a considerable effect on the perceptual appearance of the stimuli, did not qualitatively change accuracies. Quantitatively, the performance was generally higher for high coherence values (see Figure~\ref{fig:eidolon_performance} for details). Unlike in the case of contrast, the three networks showed only minor inter-network accuracy differences.

As for the contrast-experiment and the noise-experiment, we find all three networks to be strongly biased towards a few categories as shown by their low response distribution entropy (Figure~\ref{fig:accuracy_entropy}f).

\subsubsection{Performance visualization}
Here we provide a visualization of the performance differences between the studied DNNs and human observers in terms of their generalisation ability (or robustness against image degradations). For all degradation-types---contrast, noise, eidolons with different coherence parameters---we estimated the stimulus levels corresponding to $ 50\% $ classification accuracy. The stimulus levels were calculated assuming a linear relationship between the two closest data points measured in the experiments and shown in the left column of Figure~\ref{fig:accuracy_entropy}. 

\begin{figure}[t]
	\centering
	\subfloat[][Noise-experiment]{\includegraphics[width=0.5\textwidth]{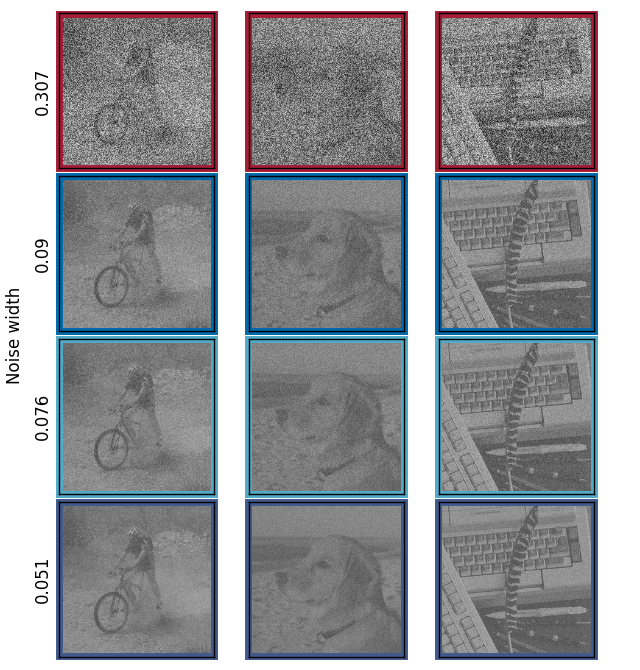}}
	\subfloat[][Eidolon-experiment (coherence = 1.0)]{\includegraphics[width=0.5\textwidth]{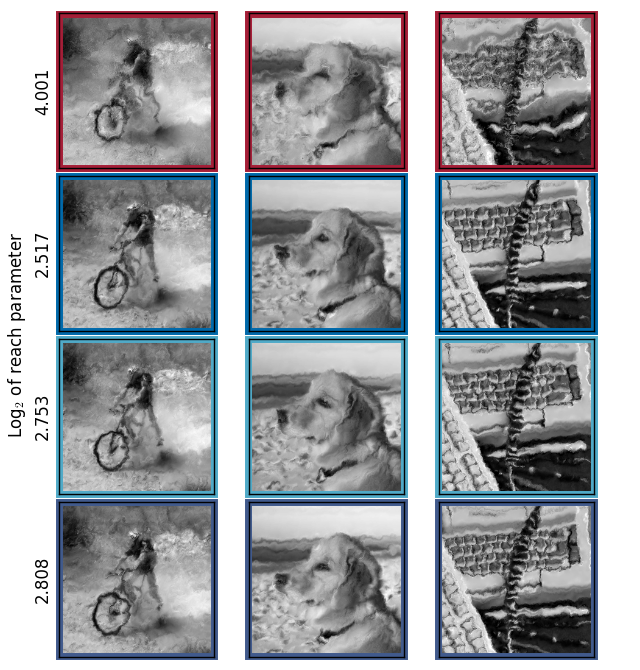}}
	\caption{Estimated stimuli corresponding to $ 50\% $ classification accuracy. \textbf{(a)} \emph{Noise-experiment}. \textbf{(b)} \emph{Eidolon-experiment}. Coherence parameter = 1.0. Top row: stimuli corresponding to threshold for the \textcolor{human.100}{average human observer}. Bottom three rows: stimuli corresponding to the same accuracy for \textcolor{vgg.100}{VGG-16} (second row), \textcolor{googlenet.100}{GoogLeNet} (third row) and \textcolor{alexnet.100}{AlexNet} (bottom row).}
	\label{fig:threshold_noise_eidolon}
\end{figure}

Figure~\ref{fig:threshold_noise_eidolon}(a) shows the $ 50\% $ accuracies for the noise-experiment, Figure~\ref{fig:threshold_noise_eidolon}(b) for the eidolon-experiment with maximal coherence (as in Figures~\ref{fig:accuracy_entropy}(e) and (f)); the three illustration images of categories \texttt{bicycle}, \texttt{dog} and \texttt{keyboard} were drawn randomly from the pool of images used in the experiments. In both panels the top row shows the stimuli corresponding to $ 50\% $ accuracy for the average human observer. The bottom three rows show the corresponding stimuli for VGG-16 (second row), GoogLeNet (third row) and AlexNet (bottom row). On a typical computer screen the more robust performance of human observers over DNNs should be readily appreciable. $ 50\% $ accuracy stimulus plots for the contrast-experiment and the other conditions of the eidolon-experiment can be found in the Appendix, Figure \ref{fig:threshold_contrast_eidolon}. 

\subsection{Confusion and confusion difference matrices}
\label{methods: confusion_difference}
\begin{figure}
	\centering
	\subfloat[Colour-experiment, colour-condition, human participants.]{
		\includegraphics[width=0.497\textwidth]{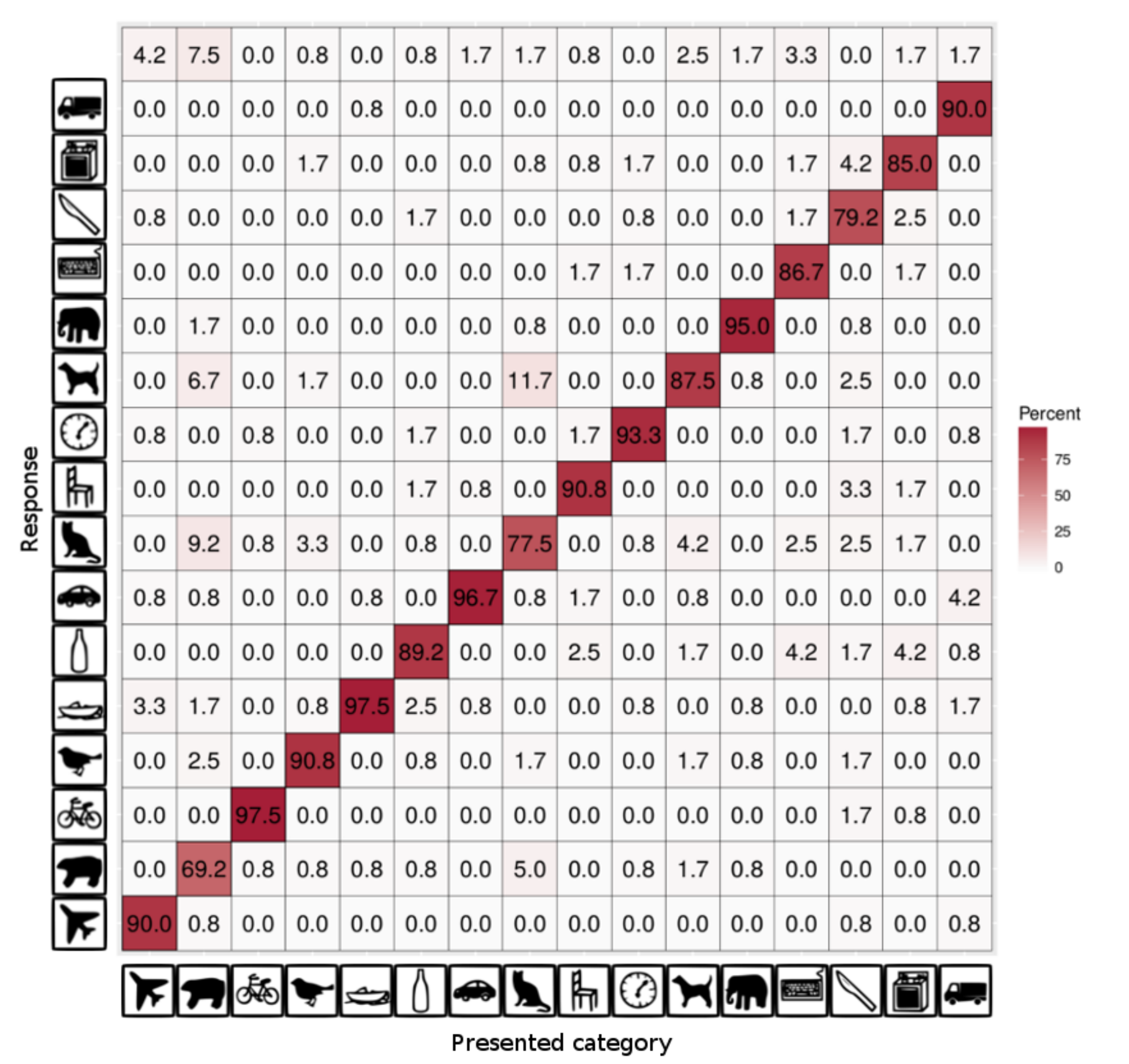}}
	\subfloat[Colour-experiment, colour-condition, difference between human participants and VGG-16.]{
	\includegraphics[width=0.50\textwidth]{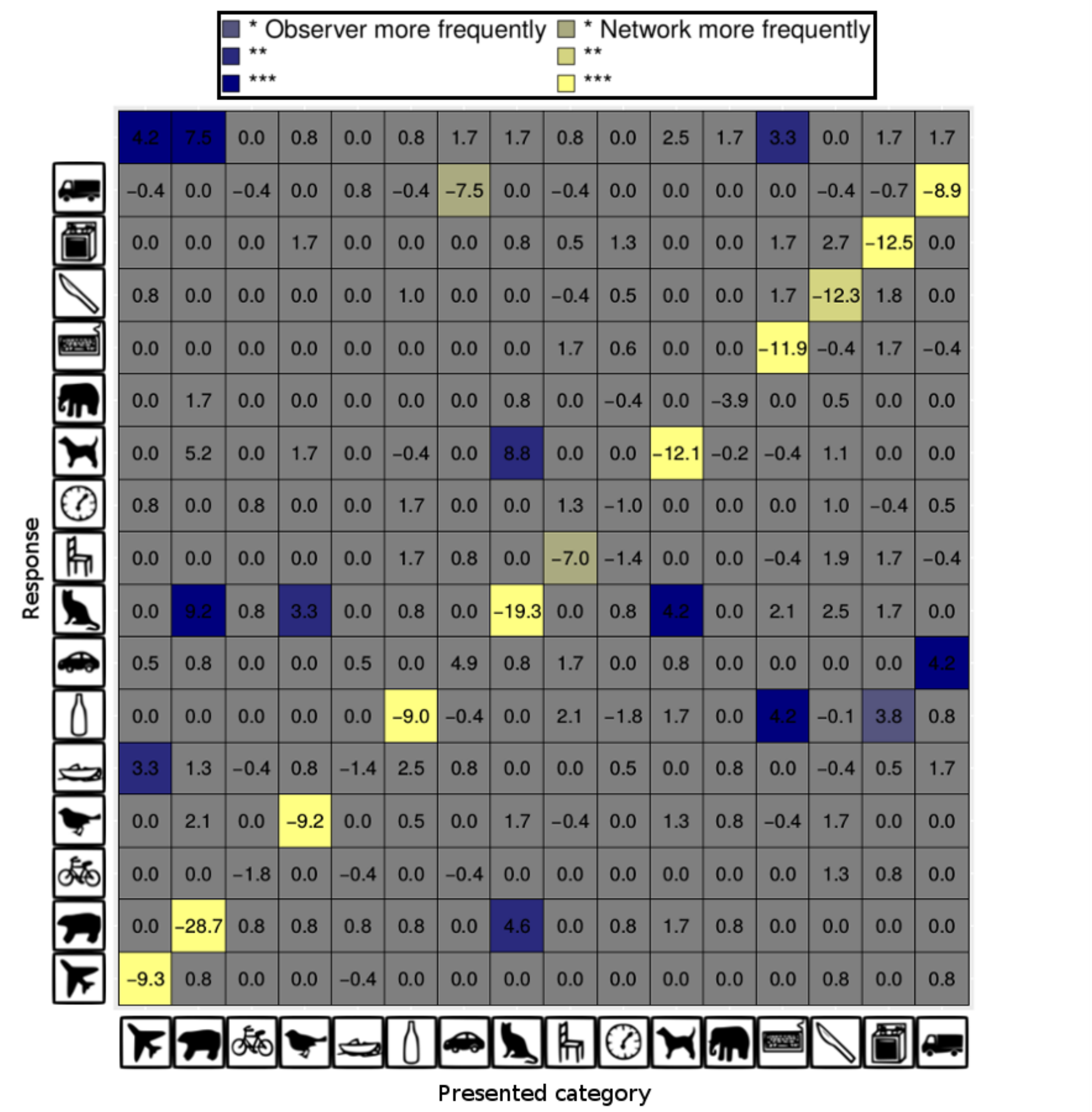}}
	\caption{Confusion and confusion difference matrices for colour-experiment (colour-condition only). A failure to respond is shown as a separate category in the top row here.
	\textbf{(a)} \emph{Standard confusion matrix}. Entries on the diagonal indicate correct classification, off-diagonal entries indicate errors.
	\textbf{(b)} \emph{Confusion difference matrix}. All values indicate the signed difference of human observers' and VGG-16's confusion matrix entries. A positive sign indicates that human observers responded more frequently than VGG-16 to a certain category-response pair, for a negative sign vice-versa. The colour here indicates whether a difference for a certain cell is significant at $ \alpha = \frac{5\%}{16\cdot 17}(*), \frac{1\%}{16\cdot 17}(**)$ and $\frac{0.1\%}{16\cdot 17}(***) $; these $ \alpha $-levels are Bonferroni corrected for multiple comparisons (16 categories $ \cdot $ 17 possible responses); see text in Section \ref{methods: confusion_difference} for details.}
	\label{fig:colour_confusion}
\end{figure}

Confusion matrices are a widely used tool for visualizing error patterns in multi-class classification data, providing insight into classification behavior (e.g.: does VGG-16 frequently confuse \texttt{dogs} with \texttt{cats}?). Figure~\ref{fig:colour_confusion}(a) shows a standard confusion matrix of the colour condition in the colour-experiment (Section~\ref{exp:colour}) for our human observers. Entries on the diagonal indicate correct classification, off-diagonal entries indicate errors, e.g. when a \texttt{cat} was presented on the screen ($8^{th}$ column from left), human observers in 77.5\% of all cases correctly clicked \texttt{cat} ($8^{th}$ row from bottom in $8^{th}$ column), but in 11.7\% clicked \texttt{dog} instead ($11^{th}$ row from bottom in $8^{th}$ column). Participants failed to respond in 1.7\% of cat trials in the colour condition of the colour-experiment (1st row from top in 8th column). Human observers typically confused physically and semantically closely related categories with each other, most notably some animal categories such as \texttt{bear, cat} and \texttt{dog}. Importantly, the same occurred for DNNs, albeit for different categories (confusions often between \texttt{car} and \texttt{truck}).

For the purpose of our analyses, however, we are mainly interested in comparisons between error patterns, e.g., do human observers more frequently confuse \texttt{dogs} with \texttt{cats} than VGG-16, and if so, significantly more? In order to being able to answer such questions, we developed a novel analysis and visualization technique, which we term \textit{confusion difference matrix}.

A confusion difference matrix serves the purpose of showing the difference between two confusion matrices (e.g. human observers and VGG-16) and highlighting which differences are significant at the indicated, Bonferroni-corrected $ \alpha $-level. A confusion difference matrix is obtained in two steps: First, one calculates the difference between two confusion matrices' entries. In this newly obtained matrix, values close to zero indicate similar classification behavior, whereas differences point at diverging classification behavior. In the next step, we calculate whether a difference for a certain cell is significant, and repeat this calculation for all cells. We calculate significance using a standard test of the probability of success in a Binomial experiment: If one thinks of the 120 colour-experiment trials in which human observers were exposed to a coloured \texttt{cat} image, of which they clicked on \texttt{cat} in 93 trials, as of a Binomial experiment with 93 successes out of 120 trials, is ``93 out of 120'' significantly higher or lower than we would expect under the null hypothesis of success probability $ p $ = 96.8\% (VGG-16's fraction of responses in this cell\footnote{It would also be possible compare VGG-16's number of successes to human observers' fraction of responses. We always compared the network/observer/group with less trials to the one with more trials as null hypothesis---in the example above (colour-experiment, colour-condition, \texttt{cat} images): a total of 120 trials for human observers vs. 280 trials for VGG-16 (or any other network).})? The Binomial tests were performed with R, using the \texttt{binom.test} function of package \texttt{stats} which calculates the conservative Clopper-Pearson confidence interval\footnote{If the network's fraction of responses in a certain cell was 0.0\% (not a single response in this cell), we set $ p = 0.1\% $ and if it was 100.0\% (every time a certain category was presented, the response lied in this cell), we set $ p = 99.9 \%$ instead.}. The significance of a certain difference, in our experiments, is not used for traditional hypothesis testing but rather as a means of distinguishing between important and unimportant---perhaps only coincidental---\emph{behavioural} differences between humans and DNNs even if their accuracies were equal. Confusion difference matrices thus visualize systematic category-dependent error pattern differences between human observers and DNNs---and they do this at a much more fine-grained, category-specific level than the response distribution entropy analyses shown in Section~\ref{results_general}.

Figure~\ref{fig:colour_confusion}(b) shows one confusion difference matrix for the colour-experiment (colour-condition only); all values indicate the signed difference of human observers' and VGG-16's confusion matrix entries. A positive sign indicates that human observers responded more frequently than VGG-16 to a certain category-response pair, for a negative sign vice-versa. VGG-16 is significantly better for many categories on the diagonal (correct classification) because---in the non-degraded colour condition---human observers make more errors, see Figure~\ref{fig:colour_performance}(a). Overall, however, most cells of the confusion difference matrix are grey, indicating very similar classification behaviour of human observers and VGG-16, not only in terms of overall accuracy and response entropy, but on a fine-grained category-by-category level.

\begin{figure}
	\centering
	\includegraphics[width=1.0\textwidth]{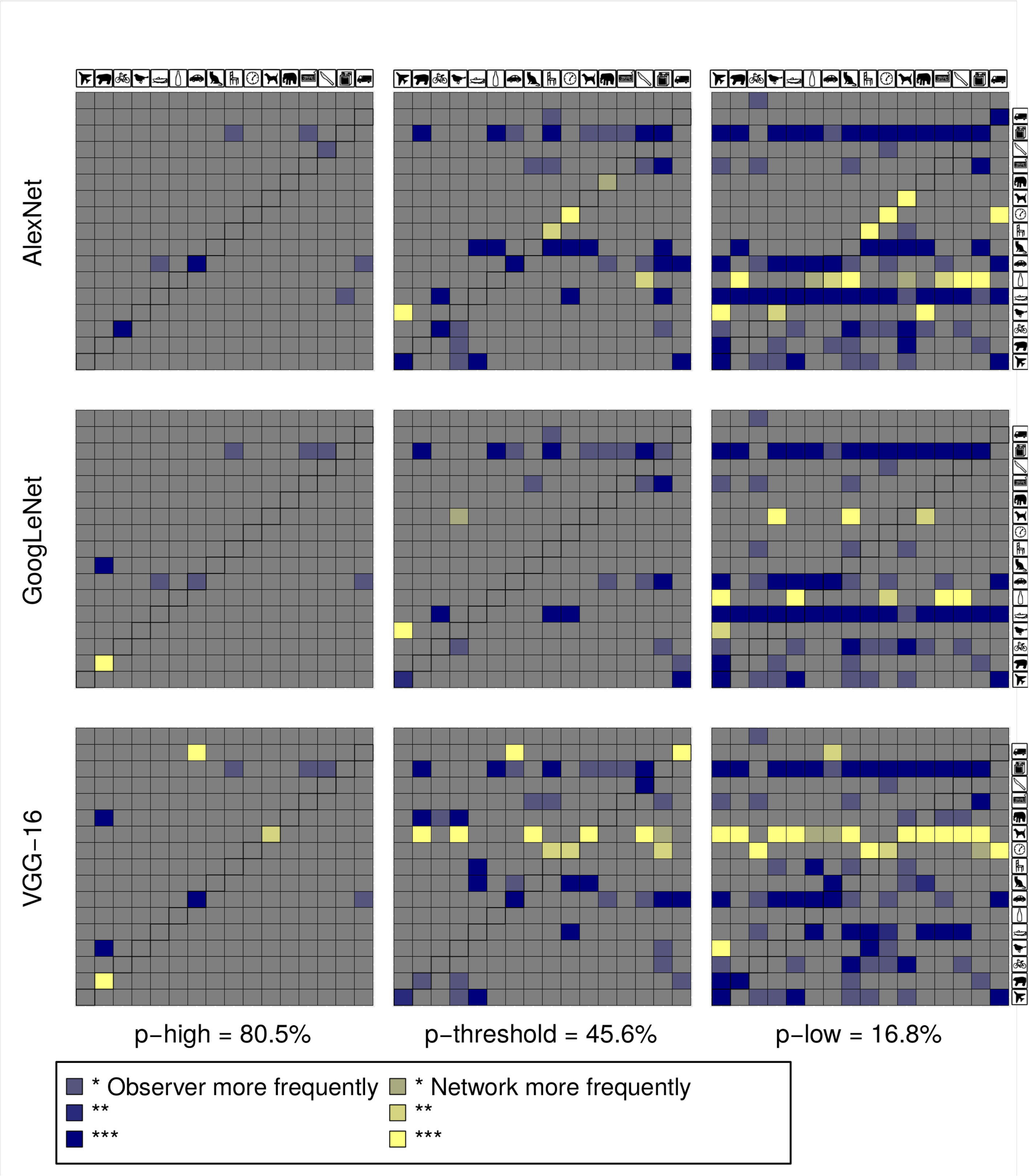}
	\caption{Confusion difference grid for the noise-experiment. Every confusion difference matrix shown here is calculated as described in the text. We here show difference matrices for which human observers and networks have similar overall performance. The left column shows for human observers the condition in which they performed best (no noise, performance p-high = 80.5\%) and for all three networks the condition for which they achieved the same accuracy. On the right, data for low performance (16.8\%) are shown and in the middle results for medium performance: 45.6\%. Colour indicates whether a difference for a certain cell is significant at $ \alpha = \frac{5\%}{16\cdot 17\cdot 9}(*), \frac{1\%}{16\cdot 17\cdot 9}(**)$ and $\frac{0.1\%}{16\cdot 17\cdot 9}(***) $; these $ \alpha $-levels are Bonferroni corrected for multiple comparisons (16 categories $ \cdot $ 17 possible responses $ \cdot $ 9 confusion difference matrices).}
	\label{fig:noise_confusion}
\end{figure}

In Figure~\ref{fig:noise_confusion} we show a confusion difference matrix grid for the noise-experiment (Section~\ref{exp:noise}): nine confusion difference matrices for all three DNNs at three matched performance levels. Confusion difference matrices shown here are calculated as described above, however, with the important difference that we here show difference matrices for which human observers and networks have similar overall performance (accuracy difference $ < 5\% $): we compare confusion matrices for different stimulus levels, but matched in performance\footnote{If DNN-human accuracy deviance was more than 5\% for all conditions, we ran additional experiments to determine a suitable condition.}. The left column shows high performance (no noise for human observers, very little noise for DNNs; performance p-high = 80.5\% which corresponds, in this order, to $w=$ 0.0, 0.0, 0.0 and 0.03 for human observers, AlexNet, GoogLeNet and VGG-16). On the right, data for low performance (16.8\%) are shown (high noise for human observers, moderate-to-low noise for DNNs, $w=$ 0.60, 0.10, 0.15, 0.19) and in the middle results for medium performance: 45.6\%, the condition for which human observers' accuracy was approximately equal to $ \frac{1}{2} $(p-high + p-low) (medium noise for human observers, low noise for DNNs; $w=$ 0.35, 0.06, 0.08, 0.10).

Showing confusion difference matrices at matched performance levels---rather than at the same stimulus level---has the advantage that the sum over all entries of the to-be-compared confusion matrices is the same, i.e. for equally behaving classifiers the expectation is to obtain mainly grey (non-significant) cells. However, inspection of Figure~\ref{fig:noise_confusion} shows this only to be the case for the easy, low-noise, condition (left column). With increasing task difficulty (more noise), network and human behavior diverges substantially. As the noise level increases, all networks show a rapidly increasing bias for a few categories. For a noise level of $ w = 0.35 $, AlexNet and GoogLeNet almost exclusively respond \texttt{bottle} ($ 92.32\% $ and $ 85.71\% $), whereas VGG-16 homes in on category \texttt{dog} for $ 62.50 \% $ of all images. Note that this bias for certain categories is neither consistent across networks nor across the image manipulations.

A similar pattern emerged for the contrast-experiment: Classification behavior on a stimulus-by-stimulus basis for all three DNNs is close to that of human observers for a high accuracy (nominal contrast level). However, as task difficulty increases, the classification behavior of all three DNNs differs significantly from human behavior, despite being matched in overall accuracy (see the Appendix, Figure~\ref{fig:contrast_confusion}).

\section{Discussion}
  \label{Discussion}
We psychophysically examined to what extend currently well-known DNNs (AlexNet, GoogLeNet and VGG-16) could be a good model for human feedforward visual object recognition. So far thorough comparisons of DNNs and human observers on behavioural grounds have been rare. Here we proposed a fair and psychophysically accurate way of comparing network and human performance on a number of object recognition tasks: measuring categorization accuracy for single-fixation, briefly presented (200 ms) and backward-masked images as a function of colour, contrast, uniform noise, and eidolon-type distortions.

We find that DNNs outperform human observers by a significant margin for non-distorted, coloured images---the images the DNNs were specifically trained on. We speculate that this may in part be due to some images in the ImageNet database containing images with small animals in the background, making it tough to decide whether it is a \texttt{cat}, \texttt{dog} or even a \texttt{bear}. Given that the images were labelled by human observers---who thus are the ultimate guardians of what counts as right or wrong---it is clear that for unlimited inspection time and sufficient training human observers will equal DNN performance, as shown by the benchmark results of \citeA{Russakovsky2015}, obtained using expert annotators. What we established, however, is that under conditions minimizing feedback, current DNNs already outperform human observers on the type of images found on ImageNet..

Our first experiment also shows that human observers' accuracy suffers only marginally when images are converted to grayscale in comparison to coloured images, consistent with previous studies \cite{Delorme2000, Kubilius2016, Wichmann2006}\footnote{Consistent, also, with the popularity of black-and-white movies and photography: If we had a hard time recognizing objects and scenes in black-and-white we doubt that they'd ever have been a mass medium in the early and mid 20th century.}. For all three tested DNNs the performance decrement is significant, however. Particularly AlexNet shows a largish drop in performance ($ > 7\% $), which is not human-like. VGG-16 and GoogLeNet rely less on colour information, but still somewhat more than the average human observer.

Our second experiment examined accuracy as a function of image contrast. Human participants outperform AlexNet and GoogLeNet (but not VGG-16) in the low contrast regime, where all DNNs display an increasing bias for certain categories (Figure~\ref{fig:accuracy_entropy}(b) as well as Figure~\ref{fig:contrast_confusion}). Almost all images on which the networks were originally trained with had full contrast. Apparently, training on ImageNet in itself only leads to a suboptimal contrast invariance. There are several solutions to overcome this deficiency: One option would be to include an explicit image preprocessing stage or have the first layer of the networks normalise the contrast. Another option would be to augment the training data with images of various contrast levels, which in itself might be a worthwhile data augmentation technique even if one does not expect low contrast images at test time. In the human visual system, probably as a response to the requirement of increasing stimulus identification accuracy \cite{Geisler_Albrecht_1995}, a mechanism called \textit{contrast gain control} evolved, serving the human visual system as a contrast normalization technique by taking into account the average local contrast rather than the absolute, global contrast \cite<e.g.>{Carandini1997, Heeger_1992, Sinz2009, Sinz2013}. This has the (side-) effect that human observers can easily perform object recognition across a variety of contrast levels. Thus yet another, though clearly more labour-intensive way of improving contrast invariance in DNNs would be to incorporate a mechanism of contrast gain control directly in the network architecture. Early vision models could serve as a role model \cite<e.g.,>{Goris2013, Schuett2016b}.

Our third experiment, adding uniform white noise to images, shows very clear discrepancies between the performance of DNNs and human observers. Note that, if anything, we might have underestimated human performance: randomly shuffling all conditions of an experiment instead of using blocks of a certain stimulus level are likely to yield accuracies that are lower than those possible in a blocked constant stimulus setting \cite{Blackwell1953, Jaekel2006}. Already at a moderate noise level, however, network accuracies drop sharply, whereas human observers are only slightly affected (visualized in Figure~\ref{fig:threshold_noise_eidolon} showing stimuli corresponding to 50\% accuracy for human observers and the three networks). Consistent with recent results by \citeA{Dodge2017}, our data clearly show that the human visual system is currently much more robust to noise than any of the investigated DNNs.

Another noteworthy finding is that the three DNNs exhibit considerable inter-model differences; their ability to cope with grayscale and different levels of contrast and noise differs substantially. In combination with other studies finding moderate to striking differences \cite<e.g.>{Cadieu2014, Kheradpisheh2016, Lake2015}, this speaks to the need of carefully distinguishing between models rather than treating DNNs as a single model type as it is perhaps sometimes done in vision science.

Recent studies on so-called \emph{adversarial examples} in DNNs have demonstrated that, for a given image, it is possible to construct a minimally perturbed version of this image which DNNs will misclassify as belonging to an arbitrary different category \cite{Szegedy2014}. Here we show that comparatively large but purely random distortions such as additive uniform noise also lead to poor network performance. 
Our detailed analyses of the network decisions offer some clues on what could contribute to robustness against these distortions, as the measurement of confusion matrices for different signal-to-noise ratios is a powerful tool to reveal important algorithmic differences of visual decision making in humans and DNNs. All three DNNs show an escalating bias towards few categories as noise power density increases (Figures~\ref{fig:accuracy_entropy} and \ref{fig:noise_confusion}), indicating that there might be something inherent to noisy images that causes the networks to select a single category. The networks might perceive the noise as being part of the object and its texture while human observers perceive the noise like a layer in front of the image (you may judge this yourself by looking at the stimuli in Figure~\ref{fig:contrast_noise_stimuli}). 
This might be the achievement of a mechanism for depth-layering of surface representations implemented by mid-level vision, which is thought to help the human brain to encode spatial relations and to order surfaces in space \cite{Kubilius2014}. 
Incorporating such a depth-layering mechanism may lead to an improvement of current DNNs, enabling them to robustly classify objects even when they are distorted in a way that the network was not exposed to during training. 
It remains subject to future investigations to determine whether such a mechanism will emerge from augmenting the training regime with different kinds of noise, or whether changes in the network architecture, potentially inspired by knowledge about mid-level vision, are necessary to achieve this feat.

One might argue that human observers, through experience and evolution, were exposed to some image distortions (e.g. fog or snow) and therefore have an advantage over current DNNs. However, an extensive exposure to eidolon-type distortions seems exceedingly unlikely. And yet, human observers were considerably better at recognising eidolon-distorted objects, largely unaffected by the different perceptual appearance for different eidolon parameter combinations (reach, coherence).
This indicates that the representations learned by the human visual system go beyond being trained on certain distortions as they generalise towards previously unseen distortions. We believe that achieving such robust representations that generalise towards novel distortions are the key to achieve robust deep neural network performance, as the number of possible distortions is literally unlimited.

\subsection{Conclusion}
We conducted a behavioural, psychophysical comparison of human and DNN object recognition robustness against image degradations. While it has long been noticed that DNNs are extremely fragile against adversarial attacks, our results show that they are also more prone to random perturbations than humans. In comparison to human observers, we find the classification performance of three currently well-known DNNs trained on ImageNet---AlexNet, GoogLeNet and VGG-16---to decline rapidly with decreasing signal-to-noise ratio under image degradations like additive noise or eidolon-type distortions. Additionally, by measuring and comparing confusion matrices we find progressively diverging patterns of classification errors between humans and DNNs with weaker signals, and considerable inter-model differences. Our results demonstrate that there are still marked differences in the way humans and current DNNs process object information. We envision that our findings and the freely available behavioural datasets may provide a new useful benchmark for improving DNN robustness and a motivation for neuroscientists to search for mechanisms in the brain that could facilitate this robustness.

\section*{Author contributions}
R.G., H.H.S. and F.A.W. designed the study; R.G. performed the network experiments with input from D.H.J.J. and H.H.S.; R.G. acquired the behavioural data with input from F.A.W.; R.G., H.H.S., J.R. M.B. and F.A.W. analysed and interpreted the data. R.G. and F.A.W. wrote the paper with significant input from H.H.S., J.R., and M.B.

\section*{Acknowledgement}
This work has been funded, in part, by the German Federal Ministry of Education and Research (BMBF) through the Bernstein Computational Neuroscience Program T\"ubingen (FKZ: 01GQ1002) as well as the German Research Foundation (DFG; Sachbeihilfe Wi 2103/4-1 and SFB 1233 on ``Robust Vision''). M.B. acknowledges support by the Centre for Integrative Neuroscience T\"ubingen (EXC 307) and by the Intelligence Advanced Research Projects Activity (IARPA) via Department of Interior/Interior Business Center (DoI/IBC) contract number D16PC00003. J.R. is funded by the BOSCH Forschungsstiftung. 

We would like to thank Tom Wallis for providing the MATLAB source code of one of his experiments, and for allowing us to use and modify it; Silke Gramer for administrative and Uli Wannek for technical support, as well as Britta Lewke for the method of creating response icons and Patricia Rubisch for help with testing human observers.

\bibliography{refs.bib}

\begin{thebibliography}{}

\bibitem [\protect \citeauthoryear {%
Atick%
}{%
Atick%
}{%
{\protect \APACyear {1992}}%
}]{%
Atick1992}
\APACinsertmetastar {%
Atick1992}%
\begin{APACrefauthors}%
Atick, J\BPBI J.%
\end{APACrefauthors}%
\unskip\
\newblock
\APACrefYearMonthDay{1992}{}{}.
\newblock
{\BBOQ}\APACrefatitle {Could information theory provide an ecological theory of
  sensory processing?} {Could information theory provide an ecological theory
  of sensory processing?}{\BBCQ}
\newblock
\APACjournalVolNumPages{Network: Computation in neural
  systems}{3}{2}{213--251}.
\PrintBackRefs{\CurrentBib}

\bibitem [\protect \citeauthoryear {%
Barlow%
}{%
Barlow%
}{%
{\protect \APACyear {1961}}%
}]{%
Barlow1961}
\APACinsertmetastar {%
Barlow1961}%
\begin{APACrefauthors}%
Barlow, H\BPBI B.%
\end{APACrefauthors}%
\unskip\
\newblock
\APACrefYearMonthDay{1961}{}{}.
\newblock
{\BBOQ}\APACrefatitle {Possible principles underlying the transformations of
  sensory messages} {Possible principles underlying the transformations of
  sensory messages}.{\BBCQ}
\newblock
\APACjournalVolNumPages{Sensory Communication}{}{}{217--234}.
\PrintBackRefs{\CurrentBib}

\bibitem [\protect \citeauthoryear {%
Biederman%
}{%
Biederman%
}{%
{\protect \APACyear {1987}}%
}]{%
Biederman_1987}
\APACinsertmetastar {%
Biederman_1987}%
\begin{APACrefauthors}%
Biederman, I.%
\end{APACrefauthors}%
\unskip\
\newblock
\APACrefYearMonthDay{1987}{}{}.
\newblock
{\BBOQ}\APACrefatitle {Recognition-by-components: a theory of human image
  understanding} {Recognition-by-components: a theory of human image
  understanding}.{\BBCQ}
\newblock
\APACjournalVolNumPages{Psychological Review}{94}{2}{115-147}.
\PrintBackRefs{\CurrentBib}

\bibitem [\protect \citeauthoryear {%
Blackwell%
}{%
Blackwell%
}{%
{\protect \APACyear {1953}}%
}]{%
Blackwell1953}
\APACinsertmetastar {%
Blackwell1953}%
\begin{APACrefauthors}%
Blackwell, H\BPBI R.%
\end{APACrefauthors}%
\unskip\
\newblock
\APACrefYearMonthDay{1953}{}{}.
\newblock
{\BBOQ}\APACrefatitle {Psychophysical thresholds: experimental studies of
  methods of measurement} {Psychophysical thresholds: experimental studies of
  methods of measurement}.{\BBCQ}
\newblock
\APACjournalVolNumPages{University of Michigan Engineering Research Institute
  Bulletin}{No. 36}{}{xiii-227}.
\PrintBackRefs{\CurrentBib}

\bibitem [\protect \citeauthoryear {%
Brainard%
}{%
Brainard%
}{%
{\protect \APACyear {1997}}%
}]{%
Brainard1997}
\APACinsertmetastar {%
Brainard1997}%
\begin{APACrefauthors}%
Brainard, D\BPBI H.%
\end{APACrefauthors}%
\unskip\
\newblock
\APACrefYearMonthDay{1997}{}{}.
\newblock
{\BBOQ}\APACrefatitle {The psychophysics toolbox} {The psychophysics
  toolbox}.{\BBCQ}
\newblock
\APACjournalVolNumPages{Spatial Vision}{10}{}{433--436}.
\PrintBackRefs{\CurrentBib}

\bibitem [\protect \citeauthoryear {%
Cadieu%
\ \protect \BOthers {.}}{%
Cadieu%
\ \protect \BOthers {.}}{%
{\protect \APACyear {2014}}%
}]{%
Cadieu2014}
\APACinsertmetastar {%
Cadieu2014}%
\begin{APACrefauthors}%
Cadieu, C\BPBI F.%
, Hong, H.%
, Yamins, D\BPBI L\BPBI K.%
, Pinto, N.%
, Ardila, D.%
, Solomon, E\BPBI A.%
\BDBL {}DiCarlo, J\BPBI J.%
\end{APACrefauthors}%
\unskip\
\newblock
\APACrefYearMonthDay{2014}{}{}.
\newblock
{\BBOQ}\APACrefatitle {Deep neural networks rival the representation of primate
  {IT} cortex for core visual object recognition} {Deep neural networks rival
  the representation of primate {IT} cortex for core visual object
  recognition}.{\BBCQ}
\newblock
\APACjournalVolNumPages{{PLoS} Computational Biology}{10}{12}{}.
\PrintBackRefs{\CurrentBib}

\bibitem [\protect \citeauthoryear {%
Carandini%
\ \BBA {} Heeger%
}{%
Carandini%
\ \BBA {} Heeger%
}{%
{\protect \APACyear {2012}}%
}]{%
Carandini2012}
\APACinsertmetastar {%
Carandini2012}%
\begin{APACrefauthors}%
Carandini, M.%
\BCBT {}\ \BBA {} Heeger, D\BPBI J.%
\end{APACrefauthors}%
\unskip\
\newblock
\APACrefYearMonthDay{2012}{}{}.
\newblock
{\BBOQ}\APACrefatitle {Normalization as a canonical neural computation}
  {Normalization as a canonical neural computation}.{\BBCQ}
\newblock
\APACjournalVolNumPages{Nature Reviews Neuroscience}{13}{1}{51--62}.
\PrintBackRefs{\CurrentBib}

\bibitem [\protect \citeauthoryear {%
Carandini%
, Heeger%
\BCBL {}\ \BBA {} Movshon%
}{%
Carandini%
\ \protect \BOthers {.}}{%
{\protect \APACyear {1997}}%
}]{%
Carandini1997}
\APACinsertmetastar {%
Carandini1997}%
\begin{APACrefauthors}%
Carandini, M.%
, Heeger, D\BPBI J.%
\BCBL {}\ \BBA {} Movshon, J\BPBI A.%
\end{APACrefauthors}%
\unskip\
\newblock
\APACrefYearMonthDay{1997}{}{}.
\newblock
{\BBOQ}\APACrefatitle {Linearity and normalization in simple cells of the
  macaque primary visual cortex} {Linearity and normalization in simple cells
  of the macaque primary visual cortex}.{\BBCQ}
\newblock
\APACjournalVolNumPages{The Journal of Neuroscience}{17}{21}{8621--8644}.
\PrintBackRefs{\CurrentBib}

\bibitem [\protect \citeauthoryear {%
Delorme%
, Richard%
\BCBL {}\ \BBA {} Fabre-Thorpe%
}{%
Delorme%
\ \protect \BOthers {.}}{%
{\protect \APACyear {2000}}%
}]{%
Delorme2000}
\APACinsertmetastar {%
Delorme2000}%
\begin{APACrefauthors}%
Delorme, A.%
, Richard, G.%
\BCBL {}\ \BBA {} Fabre-Thorpe, M.%
\end{APACrefauthors}%
\unskip\
\newblock
\APACrefYearMonthDay{2000}{}{}.
\newblock
{\BBOQ}\APACrefatitle {Ultra-rapid categorisation of natural scenes does not
  rely on colour cues: a study in monkeys and humans} {Ultra-rapid
  categorisation of natural scenes does not rely on colour cues: a study in
  monkeys and humans}.{\BBCQ}
\newblock
\APACjournalVolNumPages{Vision Research}{40}{16}{2187--2200}.
\PrintBackRefs{\CurrentBib}

\bibitem [\protect \citeauthoryear {%
DiCarlo%
, Zoccolan%
\BCBL {}\ \BBA {} Rust%
}{%
DiCarlo%
\ \protect \BOthers {.}}{%
{\protect \APACyear {2012}}%
}]{%
DiCarlo2012}
\APACinsertmetastar {%
DiCarlo2012}%
\begin{APACrefauthors}%
DiCarlo, J\BPBI J.%
, Zoccolan, D.%
\BCBL {}\ \BBA {} Rust, N\BPBI C.%
\end{APACrefauthors}%
\unskip\
\newblock
\APACrefYearMonthDay{2012}{}{}.
\newblock
{\BBOQ}\APACrefatitle {How does the brain solve visual object recognition?}
  {How does the brain solve visual object recognition?}{\BBCQ}
\newblock
\APACjournalVolNumPages{Neuron}{73}{3}{415--434}.
\PrintBackRefs{\CurrentBib}

\bibitem [\protect \citeauthoryear {%
Dodge%
\ \BBA {} Karam%
}{%
Dodge%
\ \BBA {} Karam%
}{%
{\protect \APACyear {2017}}%
}]{%
Dodge2017}
\APACinsertmetastar {%
Dodge2017}%
\begin{APACrefauthors}%
Dodge, S.%
\BCBT {}\ \BBA {} Karam, L.%
\end{APACrefauthors}%
\unskip\
\newblock
\APACrefYearMonthDay{2017}{}{}.
\newblock
\APACrefbtitle {A Study and Comparison of Human and Deep Learning Recognition
  Performance Under Visual Distortions.} {A study and comparison of human and
  deep learning recognition performance under visual distortions.}
\newblock
\APAChowpublished {arXiv preprint arXiv:1705.02498}.
\PrintBackRefs{\CurrentBib}

\bibitem [\protect \citeauthoryear {%
Douglas%
\ \BBA {} Martin%
}{%
Douglas%
\ \BBA {} Martin%
}{%
{\protect \APACyear {1991}}%
}]{%
Douglas_Martin_1991}
\APACinsertmetastar {%
Douglas_Martin_1991}%
\begin{APACrefauthors}%
Douglas, R\BPBI J.%
\BCBT {}\ \BBA {} Martin, K\BPBI A\BPBI C.%
\end{APACrefauthors}%
\unskip\
\newblock
\APACrefYearMonthDay{1991}{}{}.
\newblock
{\BBOQ}\APACrefatitle {Opening the grey box} {Opening the grey box}.{\BBCQ}
\newblock
\APACjournalVolNumPages{Trends in Neurosciences}{14}{7}{286-293}.
\PrintBackRefs{\CurrentBib}

\bibitem [\protect \citeauthoryear {%
Geirhos%
\ \protect \BOthers {.}}{%
Geirhos%
\ \protect \BOthers {.}}{%
{\protect \APACyear {2018}}%
}]{%
Geirhos2018generalisation}
\APACinsertmetastar {%
Geirhos2018generalisation}%
\begin{APACrefauthors}%
Geirhos, R.%
, Temme, C\BPBI R.%
, Rauber, J.%
, Sch{\"u}tt, H\BPBI H.%
, Bethge, M.%
\BCBL {}\ \BBA {} Wichmann, F\BPBI A.%
\end{APACrefauthors}%
\unskip\
\newblock
\APACrefYearMonthDay{2018}{}{}.
\newblock
{\BBOQ}\APACrefatitle {Generalisation in humans and deep neural networks}
  {Generalisation in humans and deep neural networks}.{\BBCQ}
\newblock
\BIn{} \APACrefbtitle {Advances in Neural Information Processing Systems}
  {Advances in neural information processing systems}\ (\BPGS\ 7548--7560).
\PrintBackRefs{\CurrentBib}

\bibitem [\protect \citeauthoryear {%
Geisler%
\ \BBA {} Albrecht%
}{%
Geisler%
\ \BBA {} Albrecht%
}{%
{\protect \APACyear {1995}}%
}]{%
Geisler_Albrecht_1995}
\APACinsertmetastar {%
Geisler_Albrecht_1995}%
\begin{APACrefauthors}%
Geisler, W\BPBI S.%
\BCBT {}\ \BBA {} Albrecht, D\BPBI G.%
\end{APACrefauthors}%
\unskip\
\newblock
\APACrefYearMonthDay{1995}{}{}.
\newblock
{\BBOQ}\APACrefatitle {Bayesian analysis of identification performance in
  monkey visual cortex: nonlinear mechanisms and stimulus certainty} {Bayesian
  analysis of identification performance in monkey visual cortex: nonlinear
  mechanisms and stimulus certainty}.{\BBCQ}
\newblock
\APACjournalVolNumPages{Vision Research}{35}{19}{2723-2730}.
\PrintBackRefs{\CurrentBib}

\bibitem [\protect \citeauthoryear {%
Gerstner%
}{%
Gerstner%
}{%
{\protect \APACyear {2005}}%
}]{%
Gerstner_2005}
\APACinsertmetastar {%
Gerstner_2005}%
\begin{APACrefauthors}%
Gerstner, W.%
\end{APACrefauthors}%
\unskip\
\newblock
\APACrefYearMonthDay{2005}{}{}.
\newblock
{\BBOQ}\APACrefatitle {How can the brain be so fast?} {How can the brain be so
  fast?}{\BBCQ}
\newblock
\BIn{} J\BPBI L.~van Hemmen\ \BBA {} T\BPBI J.~Sejnowski\ (\BEDS),
  \APACrefbtitle {23 Problems in Systems Neuroscience} {23 problems in systems
  neuroscience}\ (\BPG~135-142).
\newblock
\APACaddressPublisher{}{Oxford University Press}.
\PrintBackRefs{\CurrentBib}

\bibitem [\protect \citeauthoryear {%
Goodale%
\ \BBA {} Milner%
}{%
Goodale%
\ \BBA {} Milner%
}{%
{\protect \APACyear {1992}}%
}]{%
goodale1992}
\APACinsertmetastar {%
goodale1992}%
\begin{APACrefauthors}%
Goodale, M\BPBI A.%
\BCBT {}\ \BBA {} Milner, A\BPBI D.%
\end{APACrefauthors}%
\unskip\
\newblock
\APACrefYearMonthDay{1992}{}{}.
\newblock
{\BBOQ}\APACrefatitle {Separate visual pathways for perception and action}
  {Separate visual pathways for perception and action}.{\BBCQ}
\newblock
\APACjournalVolNumPages{Trends in Neurosciences}{15}{1}{20--25}.
\PrintBackRefs{\CurrentBib}

\bibitem [\protect \citeauthoryear {%
Goris%
, Putzeys%
, Wagemans%
\BCBL {}\ \BBA {} Wichmann%
}{%
Goris%
\ \protect \BOthers {.}}{%
{\protect \APACyear {2013}}%
}]{%
Goris2013}
\APACinsertmetastar {%
Goris2013}%
\begin{APACrefauthors}%
Goris, R\BPBI L.%
, Putzeys, T.%
, Wagemans, J.%
\BCBL {}\ \BBA {} Wichmann, F\BPBI A.%
\end{APACrefauthors}%
\unskip\
\newblock
\APACrefYearMonthDay{2013}{}{}.
\newblock
{\BBOQ}\APACrefatitle {A neural population model for visual pattern detection.}
  {A neural population model for visual pattern detection.}{\BBCQ}
\newblock
\APACjournalVolNumPages{Psychological Review}{120}{3}{472}.
\PrintBackRefs{\CurrentBib}

\bibitem [\protect \citeauthoryear {%
Green%
}{%
Green%
}{%
{\protect \APACyear {1964}}%
}]{%
Green_1964}
\APACinsertmetastar {%
Green_1964}%
\begin{APACrefauthors}%
Green, D\BPBI M.%
\end{APACrefauthors}%
\unskip\
\newblock
\APACrefYearMonthDay{1964}{}{}.
\newblock
{\BBOQ}\APACrefatitle {Consistency of Auditory Judgements} {Consistency of
  auditory judgements}.{\BBCQ}
\newblock
\APACjournalVolNumPages{Psychological Review}{71}{5}{592-407}.
\PrintBackRefs{\CurrentBib}

\bibitem [\protect \citeauthoryear {%
He%
, Zhang%
, Ren%
\BCBL {}\ \BBA {} Sun%
}{%
He%
\ \protect \BOthers {.}}{%
{\protect \APACyear {2015}}%
}]{%
He2015delving}
\APACinsertmetastar {%
He2015delving}%
\begin{APACrefauthors}%
He, K.%
, Zhang, X.%
, Ren, S.%
\BCBL {}\ \BBA {} Sun, J.%
\end{APACrefauthors}%
\unskip\
\newblock
\APACrefYearMonthDay{2015}{}{}.
\newblock
{\BBOQ}\APACrefatitle {Delving deep into rectifiers: Surpassing human-level
  performance on {ImageNet} classification} {Delving deep into rectifiers:
  Surpassing human-level performance on {ImageNet} classification}.{\BBCQ}
\newblock
\BIn{} \APACrefbtitle {{Proceedings of the IEEE International Conference on
  Computer Vision}} {{Proceedings of the IEEE International Conference on
  Computer Vision}}\ (\BPGS\ 1026--1034).
\PrintBackRefs{\CurrentBib}

\bibitem [\protect \citeauthoryear {%
Heeger%
}{%
Heeger%
}{%
{\protect \APACyear {1992}}%
}]{%
Heeger_1992}
\APACinsertmetastar {%
Heeger_1992}%
\begin{APACrefauthors}%
Heeger, D\BPBI J.%
\end{APACrefauthors}%
\unskip\
\newblock
\APACrefYearMonthDay{1992}{}{}.
\newblock
{\BBOQ}\APACrefatitle {Normalization of cell responses in cat striate cortex}
  {Normalization of cell responses in cat striate cortex}.{\BBCQ}
\newblock
\APACjournalVolNumPages{Visual Neuroscience}{9}{}{181-197}.
\PrintBackRefs{\CurrentBib}

\bibitem [\protect \citeauthoryear {%
Henning%
, Bird%
\BCBL {}\ \BBA {} Wichmann%
}{%
Henning%
\ \protect \BOthers {.}}{%
{\protect \APACyear {2002}}%
}]{%
Henning_etal_2002b}
\APACinsertmetastar {%
Henning_etal_2002b}%
\begin{APACrefauthors}%
Henning, G\BPBI B.%
, Bird, C\BPBI M.%
\BCBL {}\ \BBA {} Wichmann, F\BPBI A.%
\end{APACrefauthors}%
\unskip\
\newblock
\APACrefYearMonthDay{2002}{}{}.
\newblock
{\BBOQ}\APACrefatitle {Contrast discrimination with pulse trains in pink noise}
  {Contrast discrimination with pulse trains in pink noise}.{\BBCQ}
\newblock
\APACjournalVolNumPages{Journal of the Optical Society of America
  A}{19}{7}{1259-1266}.
\PrintBackRefs{\CurrentBib}

\bibitem [\protect \citeauthoryear {%
J{\"a}kel%
\ \BBA {} Wichmann%
}{%
J{\"a}kel%
\ \BBA {} Wichmann%
}{%
{\protect \APACyear {2006}}%
}]{%
Jaekel2006}
\APACinsertmetastar {%
Jaekel2006}%
\begin{APACrefauthors}%
J{\"a}kel, F.%
\BCBT {}\ \BBA {} Wichmann, F\BPBI A.%
\end{APACrefauthors}%
\unskip\
\newblock
\APACrefYearMonthDay{2006}{}{}.
\newblock
{\BBOQ}\APACrefatitle {Spatial four-alternative forced-choice method is the
  preferred psychophysical method for na{ï}ve observers} {Spatial
  four-alternative forced-choice method is the preferred psychophysical method
  for na{ï}ve observers}.{\BBCQ}
\newblock
\APACjournalVolNumPages{Journal of Vision}{6}{11}{1307-1322}.
\PrintBackRefs{\CurrentBib}

\bibitem [\protect \citeauthoryear {%
Jia%
\ \protect \BOthers {.}}{%
Jia%
\ \protect \BOthers {.}}{%
{\protect \APACyear {2014}}%
}]{%
Jia2014}
\APACinsertmetastar {%
Jia2014}%
\begin{APACrefauthors}%
Jia, Y.%
, Shelhamer, E.%
, Donahue, J.%
, Karayev, S.%
, Long, J.%
, Girshick, R.%
\BDBL {}Darrell, T.%
\end{APACrefauthors}%
\unskip\
\newblock
\APACrefYearMonthDay{2014}{}{}.
\newblock
{\BBOQ}\APACrefatitle {Caffe: Convolutional architecture for fast feature
  embedding} {Caffe: Convolutional architecture for fast feature
  embedding}.{\BBCQ}
\newblock
\BIn{} \APACrefbtitle {{Proceedings of the 22nd ACM International Conference on
  Multimedia}} {{Proceedings of the 22nd ACM International Conference on
  Multimedia}}\ (\BPGS\ 675--678).
\PrintBackRefs{\CurrentBib}

\bibitem [\protect \citeauthoryear {%
Kheradpisheh%
, Ghodrati%
, Ganjtabesh%
\BCBL {}\ \BBA {} Masquelier%
}{%
Kheradpisheh%
\ \protect \BOthers {.}}{%
{\protect \APACyear {2016}}%
}]{%
Kheradpisheh2016}
\APACinsertmetastar {%
Kheradpisheh2016}%
\begin{APACrefauthors}%
Kheradpisheh, S\BPBI R.%
, Ghodrati, M.%
, Ganjtabesh, M.%
\BCBL {}\ \BBA {} Masquelier, T.%
\end{APACrefauthors}%
\unskip\
\newblock
\APACrefYearMonthDay{2016}{}{}.
\newblock
\APACrefbtitle {Deep Networks Resemble Human Feed-forward Vision in Invariant
  Object Recognition.} {Deep networks resemble human feed-forward vision in
  invariant object recognition.}
\newblock
\APAChowpublished {arXiv preprint arXiv:1508.03929}.
\PrintBackRefs{\CurrentBib}

\bibitem [\protect \citeauthoryear {%
Kietzmann%
, McClure%
\BCBL {}\ \BBA {} Kriegeskorte%
}{%
Kietzmann%
\ \protect \BOthers {.}}{%
{\protect \APACyear {2017}}%
}]{%
Kietzmann_etal_2017}
\APACinsertmetastar {%
Kietzmann_etal_2017}%
\begin{APACrefauthors}%
Kietzmann, T\BPBI C.%
, McClure, P.%
\BCBL {}\ \BBA {} Kriegeskorte, N.%
\end{APACrefauthors}%
\unskip\
\newblock
\APACrefYearMonthDay{2017}{}{}.
\newblock
{\BBOQ}\APACrefatitle {Deep neural networks in computational neuroscience}
  {Deep neural networks in computational neuroscience}.{\BBCQ}
\newblock
\APACjournalVolNumPages{bioRxiv}{http://dx.doi.org/10.1101/133504}{}{}.
\PrintBackRefs{\CurrentBib}

\bibitem [\protect \citeauthoryear {%
Kleiner%
\ \protect \BOthers {.}}{%
Kleiner%
\ \protect \BOthers {.}}{%
{\protect \APACyear {2007}}%
}]{%
Kleiner2007}
\APACinsertmetastar {%
Kleiner2007}%
\begin{APACrefauthors}%
Kleiner, M.%
, Brainard, D.%
, Pelli, D.%
, Ingling, A.%
, Murray, R.%
\BCBL {}\ \BBA {} Broussard, C.%
\end{APACrefauthors}%
\unskip\
\newblock
\APACrefYearMonthDay{2007}{}{}.
\newblock
{\BBOQ}\APACrefatitle {{What’s new in Psychtoolbox-3}} {{What’s new in
  Psychtoolbox-3}}.{\BBCQ}
\newblock
\APACjournalVolNumPages{Perception}{36}{14}{1}.
\PrintBackRefs{\CurrentBib}

\bibitem [\protect \citeauthoryear {%
Koenderink%
, Valsecchi%
, van Doorn%
, Wagemans%
\BCBL {}\ \BBA {} Gegenfurtner%
}{%
Koenderink%
\ \protect \BOthers {.}}{%
{\protect \APACyear {2017}}%
}]{%
Koenderink2017}
\APACinsertmetastar {%
Koenderink2017}%
\begin{APACrefauthors}%
Koenderink, J.%
, Valsecchi, M.%
, van Doorn, A.%
, Wagemans, J.%
\BCBL {}\ \BBA {} Gegenfurtner, K.%
\end{APACrefauthors}%
\unskip\
\newblock
\APACrefYearMonthDay{2017}{}{}.
\newblock
{\BBOQ}\APACrefatitle {Eidolons: Novel stimuli for vision research} {Eidolons:
  Novel stimuli for vision research}.{\BBCQ}
\newblock
\APACjournalVolNumPages{Journal of Vision}{17}{2}{7--7}.
\PrintBackRefs{\CurrentBib}

\bibitem [\protect \citeauthoryear {%
Kriegeskorte%
}{%
Kriegeskorte%
}{%
{\protect \APACyear {2015}}%
}]{%
Kriegeskorte2015}
\APACinsertmetastar {%
Kriegeskorte2015}%
\begin{APACrefauthors}%
Kriegeskorte, N.%
\end{APACrefauthors}%
\unskip\
\newblock
\APACrefYearMonthDay{2015}{}{}.
\newblock
{\BBOQ}\APACrefatitle {Deep Neural Networks: A New Framework for Modeling
  Biological Vision and Brain Information Processing} {Deep neural networks: A
  new framework for modeling biological vision and brain information
  processing}.{\BBCQ}
\newblock
\APACjournalVolNumPages{Annual Review of Vision Science}{1}{15}{417--446}.
\PrintBackRefs{\CurrentBib}

\bibitem [\protect \citeauthoryear {%
Krizhevsky%
, Sutskever%
\BCBL {}\ \BBA {} Hinton%
}{%
Krizhevsky%
\ \protect \BOthers {.}}{%
{\protect \APACyear {2012}}%
}]{%
Krizhevsky2012}
\APACinsertmetastar {%
Krizhevsky2012}%
\begin{APACrefauthors}%
Krizhevsky, A.%
, Sutskever, I.%
\BCBL {}\ \BBA {} Hinton, G\BPBI E.%
\end{APACrefauthors}%
\unskip\
\newblock
\APACrefYearMonthDay{2012}{}{}.
\newblock
{\BBOQ}\APACrefatitle {{ImageNet} classification with deep convolutional neural
  networks} {{ImageNet} classification with deep convolutional neural
  networks}.{\BBCQ}
\newblock
\BIn{} \APACrefbtitle {{Advances in Neural Information Processing Systems}}
  {{Advances in Neural Information Processing Systems}}\ (\BPGS\ 1097--1105).
\PrintBackRefs{\CurrentBib}

\bibitem [\protect \citeauthoryear {%
Kubilius%
, Bracci%
\BCBL {}\ \BBA {} Op~de Beeck%
}{%
Kubilius%
\ \protect \BOthers {.}}{%
{\protect \APACyear {2016}}%
}]{%
Kubilius2016}
\APACinsertmetastar {%
Kubilius2016}%
\begin{APACrefauthors}%
Kubilius, J.%
, Bracci, S.%
\BCBL {}\ \BBA {} Op~de Beeck, H\BPBI P.%
\end{APACrefauthors}%
\unskip\
\newblock
\APACrefYearMonthDay{2016}{}{}.
\newblock
{\BBOQ}\APACrefatitle {Deep Neural Networks as a Computational Model for Human
  Shape Sensitivity} {Deep neural networks as a computational model for human
  shape sensitivity}.{\BBCQ}
\newblock
\APACjournalVolNumPages{{PLoS} Computational Biology}{12}{4}{e1004896}.
\PrintBackRefs{\CurrentBib}

\bibitem [\protect \citeauthoryear {%
Kubilius%
, Wagemans%
\BCBL {}\ \BBA {} Op~de Beeck%
}{%
Kubilius%
\ \protect \BOthers {.}}{%
{\protect \APACyear {2014}}%
}]{%
Kubilius2014}
\APACinsertmetastar {%
Kubilius2014}%
\begin{APACrefauthors}%
Kubilius, J.%
, Wagemans, J.%
\BCBL {}\ \BBA {} Op~de Beeck, H\BPBI P.%
\end{APACrefauthors}%
\unskip\
\newblock
\APACrefYearMonthDay{2014}{}{}.
\newblock
{\BBOQ}\APACrefatitle {A conceptual framework of computations in mid-level
  vision} {A conceptual framework of computations in mid-level vision}.{\BBCQ}
\newblock
\APACjournalVolNumPages{Frontiers in Computational Neuroscience}{8}{}{158}.
\PrintBackRefs{\CurrentBib}

\bibitem [\protect \citeauthoryear {%
Lake%
, Zaremba%
, Fergus%
\BCBL {}\ \BBA {} Gureckis%
}{%
Lake%
\ \protect \BOthers {.}}{%
{\protect \APACyear {2015}}%
}]{%
Lake2015}
\APACinsertmetastar {%
Lake2015}%
\begin{APACrefauthors}%
Lake, B\BPBI M.%
, Zaremba, W.%
, Fergus, R.%
\BCBL {}\ \BBA {} Gureckis, T\BPBI M.%
\end{APACrefauthors}%
\unskip\
\newblock
\APACrefYearMonthDay{2015}{}{}.
\newblock
{\BBOQ}\APACrefatitle {{Deep Neural Networks} Predict Category Typicality
  Ratings for Images} {{Deep Neural Networks} predict category typicality
  ratings for images}.{\BBCQ}
\newblock
\BIn{} \APACrefbtitle {{Proceedings of the 37th Annual Conference of the
  Cognitive Science Society}.} {{Proceedings of the 37th Annual Conference of
  the Cognitive Science Society}.}
\PrintBackRefs{\CurrentBib}

\bibitem [\protect \citeauthoryear {%
LeCun%
, Bengio%
\BCBL {}\ \BBA {} Hinton%
}{%
LeCun%
\ \protect \BOthers {.}}{%
{\protect \APACyear {2015}}%
}]{%
LeCun2015}
\APACinsertmetastar {%
LeCun2015}%
\begin{APACrefauthors}%
LeCun, Y.%
, Bengio, Y.%
\BCBL {}\ \BBA {} Hinton, G.%
\end{APACrefauthors}%
\unskip\
\newblock
\APACrefYearMonthDay{2015}{}{}.
\newblock
{\BBOQ}\APACrefatitle {Deep learning} {Deep learning}.{\BBCQ}
\newblock
\APACjournalVolNumPages{Nature}{521}{7553}{436--444}.
\PrintBackRefs{\CurrentBib}

\bibitem [\protect \citeauthoryear {%
Lin%
\ \protect \BOthers {.}}{%
Lin%
\ \protect \BOthers {.}}{%
{\protect \APACyear {2015}}%
}]{%
Lin2015}
\APACinsertmetastar {%
Lin2015}%
\begin{APACrefauthors}%
Lin, T\BHBI Y.%
, Maire, M.%
, Belongie, S.%
, Hays, J.%
, Perona, P.%
, Ramanan, D.%
\BDBL {}Zitnick, C\BPBI L.%
\end{APACrefauthors}%
\unskip\
\newblock
\APACrefYearMonthDay{2015}{}{}.
\newblock
{\BBOQ}\APACrefatitle {{Microsoft COCO: Common objects in context}} {{Microsoft
  COCO: Common objects in context}}.{\BBCQ}
\newblock
\BIn{} \APACrefbtitle {{European Conference on Computer Vision}} {{European
  Conference on Computer Vision}}\ (\BPGS\ 740--755).
\PrintBackRefs{\CurrentBib}

\bibitem [\protect \citeauthoryear {%
Nachmias%
\ \BBA {} Sansbury%
}{%
Nachmias%
\ \BBA {} Sansbury%
}{%
{\protect \APACyear {1974}}%
}]{%
Nachmias_Sansbury_1974}
\APACinsertmetastar {%
Nachmias_Sansbury_1974}%
\begin{APACrefauthors}%
Nachmias, J.%
\BCBT {}\ \BBA {} Sansbury, R\BPBI V.%
\end{APACrefauthors}%
\unskip\
\newblock
\APACrefYearMonthDay{1974}{}{}.
\newblock
{\BBOQ}\APACrefatitle {Grating contrast: Discrimination may be better than
  detection} {Grating contrast: Discrimination may be better than
  detection}.{\BBCQ}
\newblock
\APACjournalVolNumPages{Vision Research}{14}{10}{1039-1042}.
\PrintBackRefs{\CurrentBib}

\bibitem [\protect \citeauthoryear {%
Olshausen%
\ \BBA {} Field%
}{%
Olshausen%
\ \BBA {} Field%
}{%
{\protect \APACyear {1996}}%
}]{%
Olshausen1996}
\APACinsertmetastar {%
Olshausen1996}%
\begin{APACrefauthors}%
Olshausen, B\BPBI A.%
\BCBT {}\ \BBA {} Field, D\BPBI J.%
\end{APACrefauthors}%
\unskip\
\newblock
\APACrefYearMonthDay{1996}{}{}.
\newblock
{\BBOQ}\APACrefatitle {Emergence of simple-cell receptive field properties by
  learning a sparse code for natural images} {Emergence of simple-cell
  receptive field properties by learning a sparse code for natural
  images}.{\BBCQ}
\newblock
\APACjournalVolNumPages{Nature}{381}{6583}{607}.
\PrintBackRefs{\CurrentBib}

\bibitem [\protect \citeauthoryear {%
Pelli%
\ \BBA {} Farell%
}{%
Pelli%
\ \BBA {} Farell%
}{%
{\protect \APACyear {1999}}%
}]{%
Pelli_Farell_1999}
\APACinsertmetastar {%
Pelli_Farell_1999}%
\begin{APACrefauthors}%
Pelli, D\BPBI G.%
\BCBT {}\ \BBA {} Farell, B.%
\end{APACrefauthors}%
\unskip\
\newblock
\APACrefYearMonthDay{1999}{}{}.
\newblock
{\BBOQ}\APACrefatitle {Why Use noise?} {Why use noise?}{\BBCQ}
\newblock
\APACjournalVolNumPages{Journal of the Optical Society of America
  A}{16}{3}{647-653}.
\PrintBackRefs{\CurrentBib}

\bibitem [\protect \citeauthoryear {%
Potter%
}{%
Potter%
}{%
{\protect \APACyear {1976}}%
}]{%
Potter1976}
\APACinsertmetastar {%
Potter1976}%
\begin{APACrefauthors}%
Potter, M\BPBI C.%
\end{APACrefauthors}%
\unskip\
\newblock
\APACrefYearMonthDay{1976}{}{}.
\newblock
{\BBOQ}\APACrefatitle {Short-term conceptual memory for pictures} {Short-term
  conceptual memory for pictures}.{\BBCQ}
\newblock
\APACjournalVolNumPages{Journal of Experimental Psychology: human learning and
  memory}{2}{5}{509}.
\PrintBackRefs{\CurrentBib}

\bibitem [\protect \citeauthoryear {%
{R Core Team}%
}{%
{R Core Team}%
}{%
{\protect \APACyear {2016}}%
}]{%
RCoreteam}
\APACinsertmetastar {%
RCoreteam}%
\begin{APACrefauthors}%
{R Core Team}.%
\end{APACrefauthors}%
\unskip\
\newblock
\APACrefYearMonthDay{2016}{}{}.
\newblock
{\BBOQ}\APACrefatitle {R: A Language and Environment for Statistical Computing}
  {R: A language and environment for statistical computing}{\BBCQ}\
  [\bibcomputersoftwaremanual].
\newblock
\APACaddressPublisher{Vienna, Austria}{}.
\newblock
\begin{APACrefURL} \url{https://www.R-project.org/} \end{APACrefURL}
\PrintBackRefs{\CurrentBib}

\bibitem [\protect \citeauthoryear {%
Rosch%
}{%
Rosch%
}{%
{\protect \APACyear {1999}}%
}]{%
Rosch1999}
\APACinsertmetastar {%
Rosch1999}%
\begin{APACrefauthors}%
Rosch, E.%
\end{APACrefauthors}%
\unskip\
\newblock
\APACrefYearMonthDay{1999}{}{}.
\newblock
{\BBOQ}\APACrefatitle {Principles of categorization} {Principles of
  categorization}.{\BBCQ}
\newblock
\BIn{} E.~Margolis\ \BBA {} S.~Laurence\ (\BEDS), \APACrefbtitle {Concepts:
  core readings} {Concepts: core readings}\ (\BPGS\ 189--206).
\PrintBackRefs{\CurrentBib}

\bibitem [\protect \citeauthoryear {%
Russakovsky%
\ \protect \BOthers {.}}{%
Russakovsky%
\ \protect \BOthers {.}}{%
{\protect \APACyear {2015}}%
}]{%
Russakovsky2015}
\APACinsertmetastar {%
Russakovsky2015}%
\begin{APACrefauthors}%
Russakovsky, O.%
, Deng, J.%
, Su, H.%
, Krause, J.%
, Satheesh, S.%
, Ma, S.%
\BDBL {}Fei-Fei, L.%
\end{APACrefauthors}%
\unskip\
\newblock
\APACrefYearMonthDay{2015}{}{}.
\newblock
{\BBOQ}\APACrefatitle {{ImageNet Large Scale Visual Recognition Challenge}}
  {{ImageNet Large Scale Visual Recognition Challenge}}.{\BBCQ}
\newblock
\APACjournalVolNumPages{International Journal of Computer
  Vision}{115}{3}{211--252}.
\PrintBackRefs{\CurrentBib}

\bibitem [\protect \citeauthoryear {%
Sch{\"u}tt%
\ \BBA {} Wichmann%
}{%
Sch{\"u}tt%
\ \BBA {} Wichmann%
}{%
{\protect \APACyear {2017}}%
}]{%
Schuett2016b}
\APACinsertmetastar {%
Schuett2016b}%
\begin{APACrefauthors}%
Sch{\"u}tt, H\BPBI H.%
\BCBT {}\ \BBA {} Wichmann, F\BPBI A.%
\end{APACrefauthors}%
\unskip\
\newblock
\APACrefYearMonthDay{2017}{}{}.
\newblock
\APACrefbtitle {An Image-computable Psychophysical Spatial Vision Model} {An
  image-computable psychophysical spatial vision model}\ (\BVOL\ under review).
\PrintBackRefs{\CurrentBib}

\bibitem [\protect \citeauthoryear {%
Schönfelder%
\ \BBA {} Wichmann%
}{%
Schönfelder%
\ \BBA {} Wichmann%
}{%
{\protect \APACyear {2012}}%
}]{%
Schonfelder_Wichmann_2012}
\APACinsertmetastar {%
Schonfelder_Wichmann_2012}%
\begin{APACrefauthors}%
Schönfelder, V\BPBI H.%
\BCBT {}\ \BBA {} Wichmann, F\BPBI A.%
\end{APACrefauthors}%
\unskip\
\newblock
\APACrefYearMonthDay{2012}{}{}.
\newblock
{\BBOQ}\APACrefatitle {Sparse regularized regression identifies
  behaviorally-relevant stimulus features from psychophysical data} {Sparse
  regularized regression identifies behaviorally-relevant stimulus features
  from psychophysical data}.{\BBCQ}
\newblock
\APACjournalVolNumPages{Journal of the Acoustical Society of
  America}{131}{5}{3953-3969}.
\PrintBackRefs{\CurrentBib}

\bibitem [\protect \citeauthoryear {%
Silver%
\ \protect \BOthers {.}}{%
Silver%
\ \protect \BOthers {.}}{%
{\protect \APACyear {2016}}%
}]{%
silver2016}
\APACinsertmetastar {%
silver2016}%
\begin{APACrefauthors}%
Silver, D.%
, Huang, A.%
, Maddison, C\BPBI J.%
, Guez, A.%
, Sifre, L.%
, van~den Driessche, G.%
\BDBL {}Hassabis, D.%
\end{APACrefauthors}%
\unskip\
\newblock
\APACrefYearMonthDay{2016}{}{}.
\newblock
{\BBOQ}\APACrefatitle {Mastering the game of {Go} with deep neural networks and
  tree search} {Mastering the game of {Go} with deep neural networks and tree
  search}.{\BBCQ}
\newblock
\APACjournalVolNumPages{Nature}{529}{7587}{484--489}.
\PrintBackRefs{\CurrentBib}

\bibitem [\protect \citeauthoryear {%
Simoncelli%
\ \BBA {} Freeman%
}{%
Simoncelli%
\ \BBA {} Freeman%
}{%
{\protect \APACyear {1995}}%
}]{%
Simoncelli_Freeman_1995}
\APACinsertmetastar {%
Simoncelli_Freeman_1995}%
\begin{APACrefauthors}%
Simoncelli, E\BPBI P.%
\BCBT {}\ \BBA {} Freeman, W\BPBI T.%
\end{APACrefauthors}%
\unskip\
\newblock
\APACrefYearMonthDay{1995}{}{}.
\newblock
{\BBOQ}\APACrefatitle {The steerable pyramid: a flexible architecture for
  multi-scale derivative computation} {The steerable pyramid: a flexible
  architecture for multi-scale derivative computation}.{\BBCQ}
\newblock
\BIn{} \APACrefbtitle {2nd IEEE International Conference on Image Processing}
  {2nd ieee international conference on image processing}\ (\BVOL~III,
  \BPG~444-447).
\newblock
\APACaddressPublisher{Washington, DC}{}.
\PrintBackRefs{\CurrentBib}

\bibitem [\protect \citeauthoryear {%
Simoncelli%
, Freeman%
, Adelson%
\BCBL {}\ \BBA {} Heeger%
}{%
Simoncelli%
\ \protect \BOthers {.}}{%
{\protect \APACyear {1992}}%
}]{%
Simoncelli_etal_1992}
\APACinsertmetastar {%
Simoncelli_etal_1992}%
\begin{APACrefauthors}%
Simoncelli, E\BPBI P.%
, Freeman, W\BPBI T.%
, Adelson, E\BPBI H.%
\BCBL {}\ \BBA {} Heeger, D\BPBI J.%
\end{APACrefauthors}%
\unskip\
\newblock
\APACrefYearMonthDay{1992}{}{}.
\newblock
{\BBOQ}\APACrefatitle {Shiftable multiscale transforms} {Shiftable multiscale
  transforms}.{\BBCQ}
\newblock
\APACjournalVolNumPages{IEEE Transactions on Information
  Theory}{38}{2}{587-607}.
\PrintBackRefs{\CurrentBib}

\bibitem [\protect \citeauthoryear {%
Simonyan%
\ \BBA {} Zisserman%
}{%
Simonyan%
\ \BBA {} Zisserman%
}{%
{\protect \APACyear {2015}}%
}]{%
Simonyan2015}
\APACinsertmetastar {%
Simonyan2015}%
\begin{APACrefauthors}%
Simonyan, K.%
\BCBT {}\ \BBA {} Zisserman, A.%
\end{APACrefauthors}%
\unskip\
\newblock
\APACrefYearMonthDay{2015}{}{}.
\newblock
\APACrefbtitle {Very deep convolutional networks for large-scale image
  recognition.} {Very deep convolutional networks for large-scale image
  recognition.}
\newblock
\APAChowpublished {arXiv preprint arXiv:1409.1556}.
\PrintBackRefs{\CurrentBib}

\bibitem [\protect \citeauthoryear {%
Sinz%
\ \BBA {} Bethge%
}{%
Sinz%
\ \BBA {} Bethge%
}{%
{\protect \APACyear {2009}}%
}]{%
Sinz2009}
\APACinsertmetastar {%
Sinz2009}%
\begin{APACrefauthors}%
Sinz, F.%
\BCBT {}\ \BBA {} Bethge, M.%
\end{APACrefauthors}%
\unskip\
\newblock
\APACrefYearMonthDay{2009}{}{}.
\newblock
{\BBOQ}\APACrefatitle {The conjoint effect of divisive normalization and
  orientation selectivity on redundancy reduction} {The conjoint effect of
  divisive normalization and orientation selectivity on redundancy
  reduction}.{\BBCQ}
\newblock
\BIn{} \APACrefbtitle {Advances in neural information processing systems}
  {Advances in neural information processing systems}\ (\BPGS\ 1521--1528).
\PrintBackRefs{\CurrentBib}

\bibitem [\protect \citeauthoryear {%
Sinz%
\ \BBA {} Bethge%
}{%
Sinz%
\ \BBA {} Bethge%
}{%
{\protect \APACyear {2013}}%
}]{%
Sinz2013}
\APACinsertmetastar {%
Sinz2013}%
\begin{APACrefauthors}%
Sinz, F.%
\BCBT {}\ \BBA {} Bethge, M.%
\end{APACrefauthors}%
\unskip\
\newblock
\APACrefYearMonthDay{2013}{}{}.
\newblock
{\BBOQ}\APACrefatitle {Temporal adaptation enhances efficient contrast gain
  control on natural images} {Temporal adaptation enhances efficient contrast
  gain control on natural images}.{\BBCQ}
\newblock
\APACjournalVolNumPages{PLoS Comput Biol}{9}{1}{e1002889}.
\PrintBackRefs{\CurrentBib}

\bibitem [\protect \citeauthoryear {%
Szegedy%
\ \protect \BOthers {.}}{%
Szegedy%
\ \protect \BOthers {.}}{%
{\protect \APACyear {2015}}%
}]{%
Szegedy2015}
\APACinsertmetastar {%
Szegedy2015}%
\begin{APACrefauthors}%
Szegedy, C.%
, Liu, W.%
, Jia, Y.%
, Sermanet, P.%
, Reed, S.%
, Anguelov, D.%
\BDBL {}Rabinovich, A.%
\end{APACrefauthors}%
\unskip\
\newblock
\APACrefYearMonthDay{2015}{}{}.
\newblock
{\BBOQ}\APACrefatitle {Going deeper with convolutions} {Going deeper with
  convolutions}.{\BBCQ}
\newblock
\BIn{} \APACrefbtitle {Proceedings of the {IEEE Conference on Computer Vision
  and Pattern Recognition}} {Proceedings of the {IEEE Conference on Computer
  Vision and Pattern Recognition}}\ (\BPGS\ 1--9).
\PrintBackRefs{\CurrentBib}

\bibitem [\protect \citeauthoryear {%
Szegedy%
\ \protect \BOthers {.}}{%
Szegedy%
\ \protect \BOthers {.}}{%
{\protect \APACyear {2014}}%
}]{%
Szegedy2014}
\APACinsertmetastar {%
Szegedy2014}%
\begin{APACrefauthors}%
Szegedy, C.%
, Zaremba, W.%
, Sutskever, I.%
, Bruna, J.%
, Erhan, D.%
, Goodfellow, I.%
\BCBL {}\ \BBA {} Fergus, R.%
\end{APACrefauthors}%
\unskip\
\newblock
\APACrefYearMonthDay{2014}{}{}.
\newblock
\APACrefbtitle {Intriguing properties of neural networks.} {Intriguing
  properties of neural networks.}
\newblock
\APAChowpublished {arXiv preprint arXiv:1312.6199}.
\PrintBackRefs{\CurrentBib}

\bibitem [\protect \citeauthoryear {%
Thorpe%
, Fize%
\BCBL {}\ \BBA {} Marlot%
}{%
Thorpe%
\ \protect \BOthers {.}}{%
{\protect \APACyear {1996}}%
}]{%
Thorpe1996}
\APACinsertmetastar {%
Thorpe1996}%
\begin{APACrefauthors}%
Thorpe, S.%
, Fize, D.%
\BCBL {}\ \BBA {} Marlot, C.%
\end{APACrefauthors}%
\unskip\
\newblock
\APACrefYearMonthDay{1996}{}{}.
\newblock
{\BBOQ}\APACrefatitle {Speed of processing in the human visual system} {Speed
  of processing in the human visual system}.{\BBCQ}
\newblock
\APACjournalVolNumPages{Nature}{381}{6582}{520--522}.
\PrintBackRefs{\CurrentBib}

\bibitem [\protect \citeauthoryear {%
Van~der Walt%
\ \protect \BOthers {.}}{%
Van~der Walt%
\ \protect \BOthers {.}}{%
{\protect \APACyear {2014}}%
}]{%
VanderWalt2014}
\APACinsertmetastar {%
VanderWalt2014}%
\begin{APACrefauthors}%
Van~der Walt, S.%
, Sch{\"o}nberger, J\BPBI L.%
, Nunez-Iglesias, J.%
, Boulogne, F.%
, Warner, J\BPBI D.%
, Yager, N.%
\BDBL {}Yu, T.%
\end{APACrefauthors}%
\unskip\
\newblock
\APACrefYearMonthDay{2014}{}{}.
\newblock
{\BBOQ}\APACrefatitle {{scikit-image: image processing in Python}}
  {{scikit-image: image processing in Python}}.{\BBCQ}
\newblock
\APACjournalVolNumPages{PeerJ}{2}{}{e453}.
\PrintBackRefs{\CurrentBib}

\bibitem [\protect \citeauthoryear {%
VanRullen%
}{%
VanRullen%
}{%
{\protect \APACyear {2017}}%
}]{%
VanRullen_2017}
\APACinsertmetastar {%
VanRullen_2017}%
\begin{APACrefauthors}%
VanRullen, R.%
\end{APACrefauthors}%
\unskip\
\newblock
\APACrefYearMonthDay{2017}{}{}.
\newblock
{\BBOQ}\APACrefatitle {Perception Science in the Age of Deep Neural Networks}
  {Perception science in the age of deep neural networks}.{\BBCQ}
\newblock
\APACjournalVolNumPages{Frontiers in Psychology}{8}{}{142 , doi:
  10.3389/fpsyg.2017.00142}.
\PrintBackRefs{\CurrentBib}

\bibitem [\protect \citeauthoryear {%
Wichmann%
}{%
Wichmann%
}{%
{\protect \APACyear {1999}}%
}]{%
Wichmann_1999}
\APACinsertmetastar {%
Wichmann_1999}%
\begin{APACrefauthors}%
Wichmann, F\BPBI A.%
\end{APACrefauthors}%
\unskip\
\newblock
\APACrefYear{1999}.
\unskip\
\newblock
\APACrefbtitle {Some Aspects of Modelling Human Spatial Vision: Contrast
  Discrimination} {Some aspects of modelling human spatial vision: Contrast
  discrimination}\ \APACtypeAddressSchool {\BUPhD}{}{}.
\unskip\
\newblock
\APACaddressSchool {}{The University of Oxford}.
\PrintBackRefs{\CurrentBib}

\bibitem [\protect \citeauthoryear {%
Wichmann%
, Braun%
\BCBL {}\ \BBA {} Gegenfurtner%
}{%
Wichmann%
\ \protect \BOthers {.}}{%
{\protect \APACyear {2006}}%
}]{%
Wichmann2006}
\APACinsertmetastar {%
Wichmann2006}%
\begin{APACrefauthors}%
Wichmann, F\BPBI A.%
, Braun, D\BPBI I.%
\BCBL {}\ \BBA {} Gegenfurtner, K\BPBI R.%
\end{APACrefauthors}%
\unskip\
\newblock
\APACrefYearMonthDay{2006}{}{}.
\newblock
{\BBOQ}\APACrefatitle {Phase noise and the classification of natural images}
  {Phase noise and the classification of natural images}.{\BBCQ}
\newblock
\APACjournalVolNumPages{Vision Research}{46}{8}{1520--1529}.
\PrintBackRefs{\CurrentBib}

\bibitem [\protect \citeauthoryear {%
Wichmann%
\ \protect \BOthers {.}}{%
Wichmann%
\ \protect \BOthers {.}}{%
{\protect \APACyear {2017}}%
}]{%
Wichmann2017}
\APACinsertmetastar {%
Wichmann2017}%
\begin{APACrefauthors}%
Wichmann, F\BPBI A.%
, Janssen, D\BPBI H.%
, Geirhos, R.%
, Aguilar, G.%
, Sch{\"u}tt, H\BPBI H.%
, Maertens, M.%
\BCBL {}\ \BBA {} Bethge, M.%
\end{APACrefauthors}%
\unskip\
\newblock
\APACrefYearMonthDay{2017}{}{}.
\newblock
{\BBOQ}\APACrefatitle {Methods and measurements to compare men against
  machines} {Methods and measurements to compare men against machines}.{\BBCQ}
\newblock
\APACjournalVolNumPages{Electronic Imaging, Human Vision and Electronic
  Imaging}{2017}{14}{36--45}.
\PrintBackRefs{\CurrentBib}

\end{thebibliography}
\bibliographystyle{apacite}

\newpage
\renewcommand{\thesubsection}{\Alph{subsection}}
\section*{Appendix}
\appendix

\subsection{Stimuli \& Apparatus}

\subsubsection{Categories and image database}\label{methods:categories}
The images serving as psychophysical stimuli were images extracted from the training set of the ImageNet Large Scale Visual Recognition Challenge 2012 database \cite{Russakovsky2015}. This database contains millions of labeled images grouped into 1,000 very fine-grained categories (e.g., the database contains over a hundred different dog breeds). If human observers are asked to name objects, however, they most naturally categorize them into many fewer so-called basic or entry-level categories, e.g. \texttt{dog} rather than \texttt{German shepherd} \cite{Rosch1999}. The Microsoft COCO (MS COCO) database \cite{Lin2015} is an image database structured according to 91 such entry-level categories, making it an excellent source of categories for an object recognition task. Thus for our experiments we fused the carefully selected entry-level categories in the MS COCO database with the large quantity of images in ImageNet. Using WordNet's \emph{hypernym} relationship (\emph{x} is a hypernym of \emph{y} if \emph{y} is a "kind of" \emph{x}, e.g., \texttt{dog} is a hypernym of \texttt{German shepherd}), we mapped every ImageNet label to an entry-level category of MS COCO in case such a relationship exists, retaining 16 clearly non-ambiguous categories with sufficiently many images within each category (see Figure~\ref{fig:typical_trial} for a iconic representation of the 16 categories; the figure shows the icons used for the observers during the experiment). A complete list of ImageNet labels used for the experiments can be found in our github repository, \url{https://github.com/rgeirhos/object-recognition}. Since all investigated DNNs, when shown an image, output classification predictions for all 1,000 ImageNet categories, we disregarded all predictions for categories that were not mapped to any of the 16 entry-level categories. Amongst the remaining categories, the entry-level category corresponding to the ImageNet category with the highest probability (top-1) was selected as the network's response. This way, the DNN response selection corresponds directly to the forced-choice paradigm for our human observers.

\subsubsection{Image preprocessing}\label{methods: image_preprocessing}
We used Python (Version 2.7.11) for all image preprocessing and for running the DNN experiments. From the pool of ImageNet images of the 16 entry-level categories, we excluded all grayscale images (1\%) as well as all images not at least $ 256 \times 256 $ pixels in size (11\% of non-grayscale images). We then cropped all images to a center patch of $ 256 \times 256 $ pixels as follows: First, every image was cropped to the largest possible center square. This center square was then downsampled to the desired size with \texttt{PIL.Image.thumbnail((256, 256), Image.ANTIALIAS)}. Human observers get adapted to the mean luminance of the display during experiments, and thus images which are either very bright or very dark may be harder to recognize due to their very different perceived brightness. We therefore excluded all images which had a mean deviating more than two standard deviations from that of other images (5\% of correct-sized colour-images excluded). In total we retained 213,555 images from ImageNet.

\begin{figure}
	\centering
	\subfloat[][Contrast-experiment stimuli]{\includegraphics[width=0.5\textwidth]{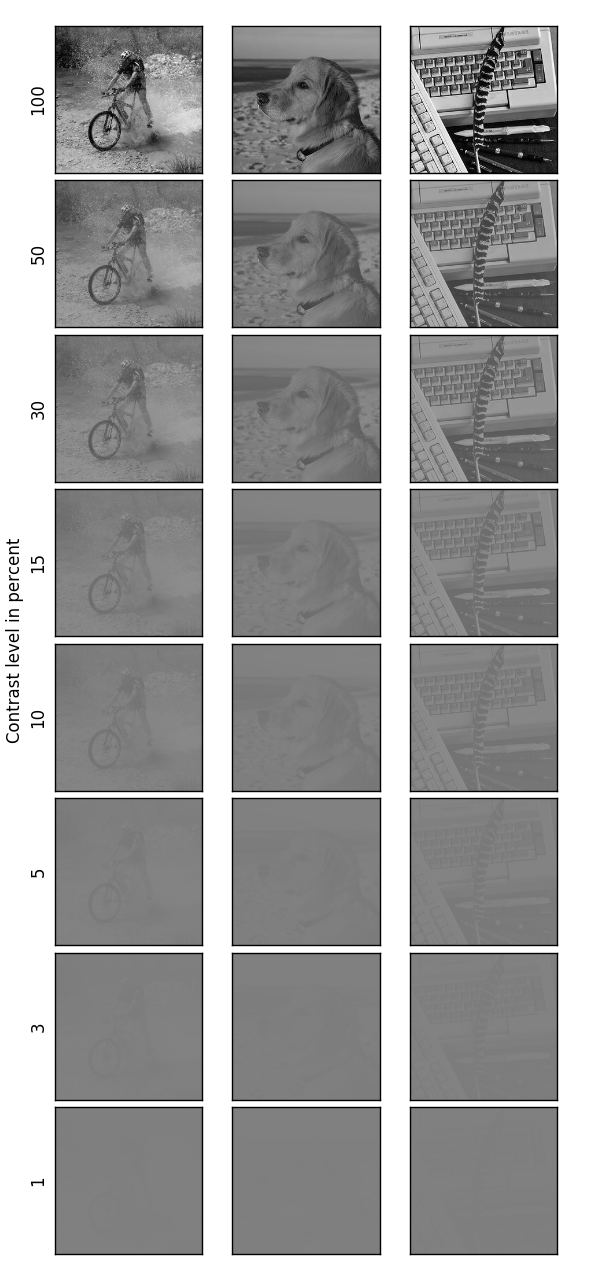}}
	\subfloat[][Noise-experiment stimuli]{\includegraphics[width=0.5\textwidth]{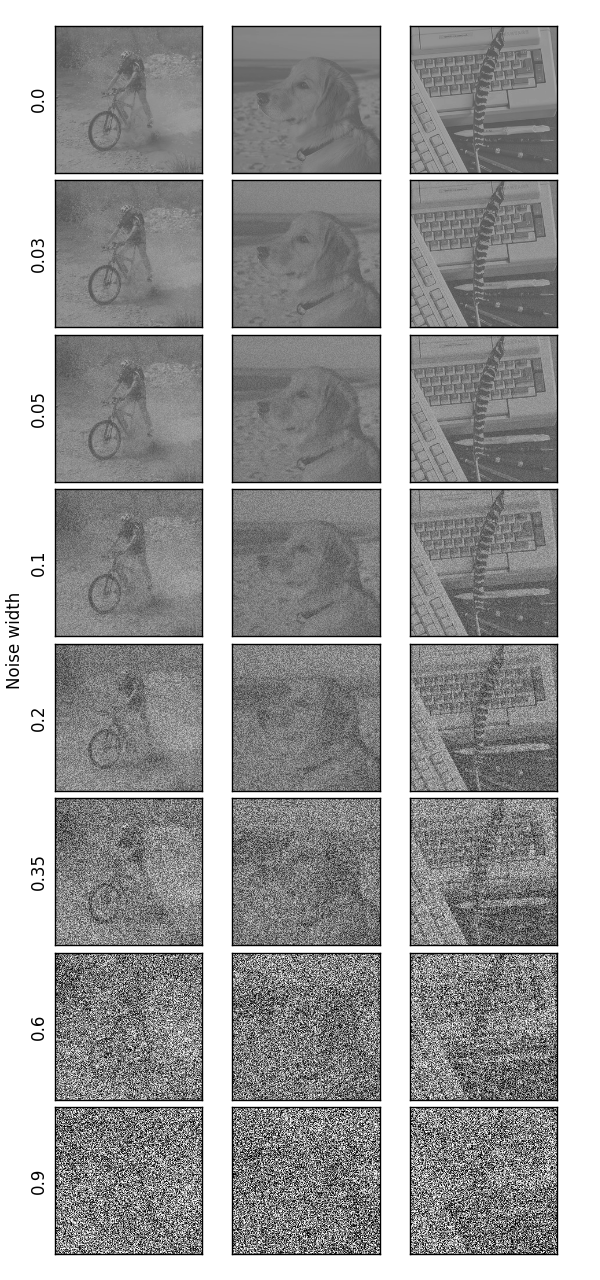}}
	\caption{Three example stimuli for all conditions of contrast-experiment and noise-experiment. The three images (categories \texttt{bicycle}, \texttt{dog} and \texttt{keyboard}) were drawn randomly from the pool of images used in the experiments.}
	\label{fig:contrast_noise_stimuli}
\end{figure}

For the experiments using grayscale images the stimuli were converted using the \texttt{rgb2gray} method \cite{VanderWalt2014} in Python. This was the case for all experiments and conditions except for the colour-condition of the colour-experiment. For the contrast-experiment, we employed eight different contrast levels $ c \in \{1, 3, 5, 10, 15, 30, 50, 100\%\}$. For an image in the [0, 1] range, scaling the image to a new contrast level $ c $ was achieved by computing $ new\_value = \frac{c}{100\%} \cdot old\_value + \frac{1-\frac{c}{100\%}}{2} $ for each pixel. For the noise-experiment, we first scaled all images to a contrast level of $ c=30\% $. Subsequently, white uniform noise of range $ [-w, w] $ was added pixelwise, $ w \in \{0.0, 0.03, 0.05, 0.1, 0.2, 0.35, 0.6, 0.9\} $. In case this resulted in a value out of the [0, 1] range, this value was clipped to either 0 or 1. By design, this never occurred for a noise range less or equal to 0.35 due to the reduced contrast (see above). For $ w = 0.6 $, clipping occurred in $ 17.2 \% $ of all pixels and for $ w = 0.9 $ in $ 44.4 \% $ of all pixels. Clearly, clipping pixels changes the spectrum of the noise and is undesirable. However, as can be seen in Section~\ref{results}, specifically Figure~\ref{fig:accuracy_entropy}, all DNNs were already at chance performance for noise with a $ w $ of 0.35 (no clipping), whereas human observers were still supra-threshold. Thus changes in the exact shape of the spectrum of the noise due to clipping have no effect on the conclusions drawn from our experiment. See Figure \ref{fig:contrast_noise_stimuli} for example contrast and noise stimuli.

\begin{figure}
	\centering
	\subfloat[][Coherence parameter = 1.0]{\includegraphics[width=0.5\textwidth]{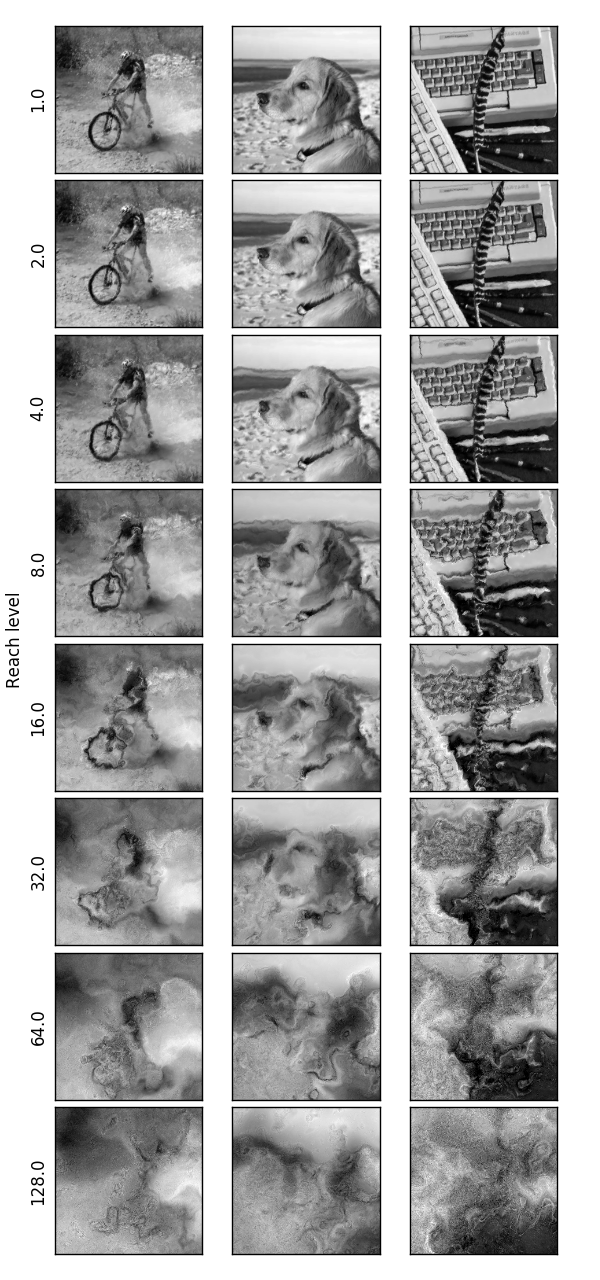}}
	\subfloat[][Coherence parameter = 0.0]{\includegraphics[width=0.5\textwidth]{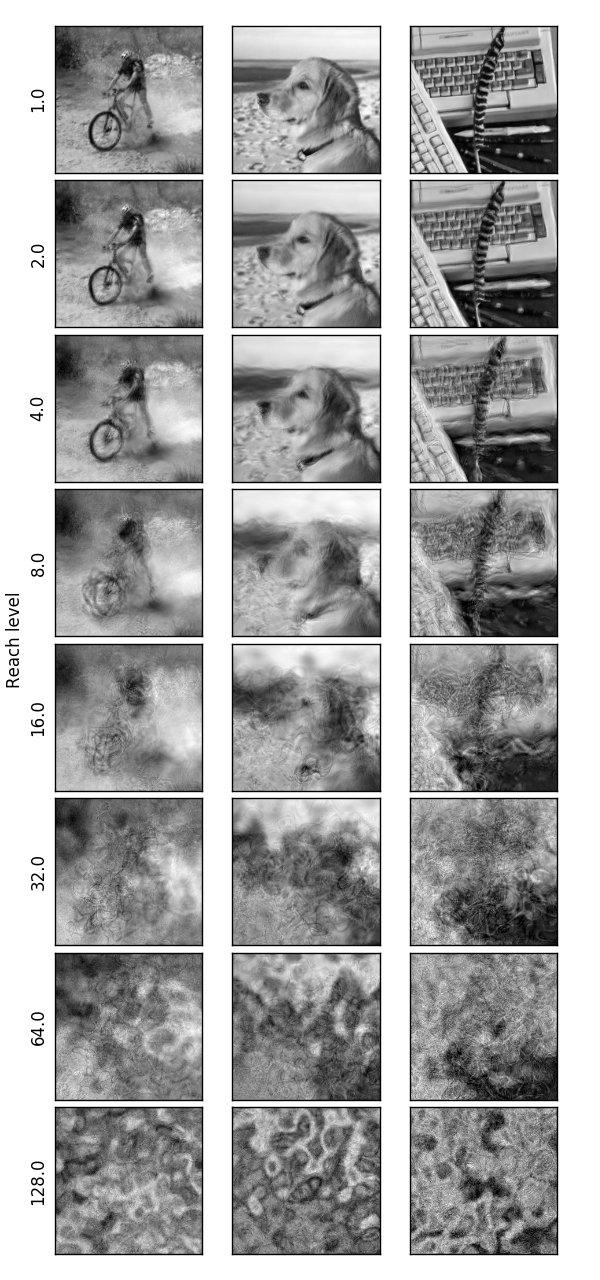}}
	\caption{Three example stimuli (\texttt{bicycle, dog, keyboard}) for all except the coherence = 0.3 conditions of the eidolon-experiment, split by coherence levels.}
	\label{fig:eidolon_stimuli}
\end{figure}

All eidolon stimuli were generated using the eidolon toolbox for Python obtained from \url{https://github.com/gestaltrevision/Eidolon}, more specifically its \texttt{PartiallyCoherentDisarray(image, reach, coherence, grain)} function.\\
Using a combination of the three parameters reach, coherence and grain, one obtains a distorted version of the original image (a so-called eidolon). The parameters reach and coherence were varied in the experiment, grain was held constant with a value of 10.0 throughout the experiment (grain indicates how fine-grained the distortion is; a value of 10.0 corresponds to a medium-grainy distortion). Reach $ \in \{1.0, 2.0, 4.0, 8.0, 16.0, 32.0, 64.0, 128.0\} $ is an amplitude-like parameter indicating the strength of the distortion, coherence $ \in \{0.0, 0.3, 1.0\}$ defines the relationship between local and global image structure. Those two parameters were fully crossed, resulting in $ 8 \cdot 3 = 24 $ different eidolon conditions. A high coherence value "retains the local image structure even when the global image structure is destroyed" \cite[p. 10]{Koenderink2017}. A coherence value of 0.0 corresponds to 'completely incoherent', a value of 1.0 to 'fully coherent'. The third value 0.3 was chosen because it produces images that perceptually lie---as informally determined by the authors---in the middle between those two extremes. See Figure~\ref{fig:eidolon_stimuli} for example eidolon stimuli.

All images, prior to showing them to human observers or DNNs, were saved in the JPEG format using the default settings of the \texttt{skimage.io.imsave} function. The JPEG format was chosen because the image training database for all three networks, ImageNet \cite{Russakovsky2015}, consists of JPEG images. However, one has to bear in mind that JPEG compression is lossy and introduces, under certain circumstances, unwanted artefacts. We therefore ran all DNN experiments additionally saving them in the (up to rounding issues) lossless PNG format. We did not find any noteworthy differences in DNN results for colour-, noise- and eidolon-experiment but did find some for the contrast-experiment, which is why we report data for PNG images in the case of the contrast-experiment (Figure \ref{fig:accuracy_entropy}). In particular, saving a low-contrast image to JPEG may result in a slightly different contrast level, which is why we refer to the contrast level of JPEG images as \textit{nominal contrast} throughout this paper. For an in-depth overview about JPEG vs. PNG results, see Section~\ref{appendix: jpeg_vs_png} of this Appendix.

\clearpage

\subsubsection{Apparatus}\label{methods: apparatus}
All stimuli were presented on a VIEWPixx LCD monitor (VPixx Technologies, Saint-Bruno, Canada) in a dark chamber. The 22'' monitor ($ 484\times 302$ mm) had a spatial resolution of  $1920 \times 1200$  pixels at a refresh rate of 120 Hz. Stimuli were presented at the center of the screen with $256 \times 256$ pixels, corresponding, at a viewing distance of 123 cm, to $3 \times 3$ degrees of visual angle. A chin rest was used in order to keep the position of the head constant over the course of an experiment. Stimulus presentation and response recording were controlled using MATLAB (Release 2016a, The MathWorks, Inc., Natick, Massachusetts, United States) and the Psychophysics Toolbox extensions version 3.0.12 \cite{Brainard1997, Kleiner2007} along with our in-house iShow library (\url{http://dx.doi.org/10.5281/zenodo.34217}) on a desktop computer (12 core CPU i7-3930K, AMD HD7970 graphics card “Tahiti” by AMD, Sunnyvale, California, United States) running Kubuntu 14.04 LTS. Responses were collected with a standard computer mouse.

\begin{figure}[h!]
	\centering
	\includegraphics[width=1.0\textwidth]{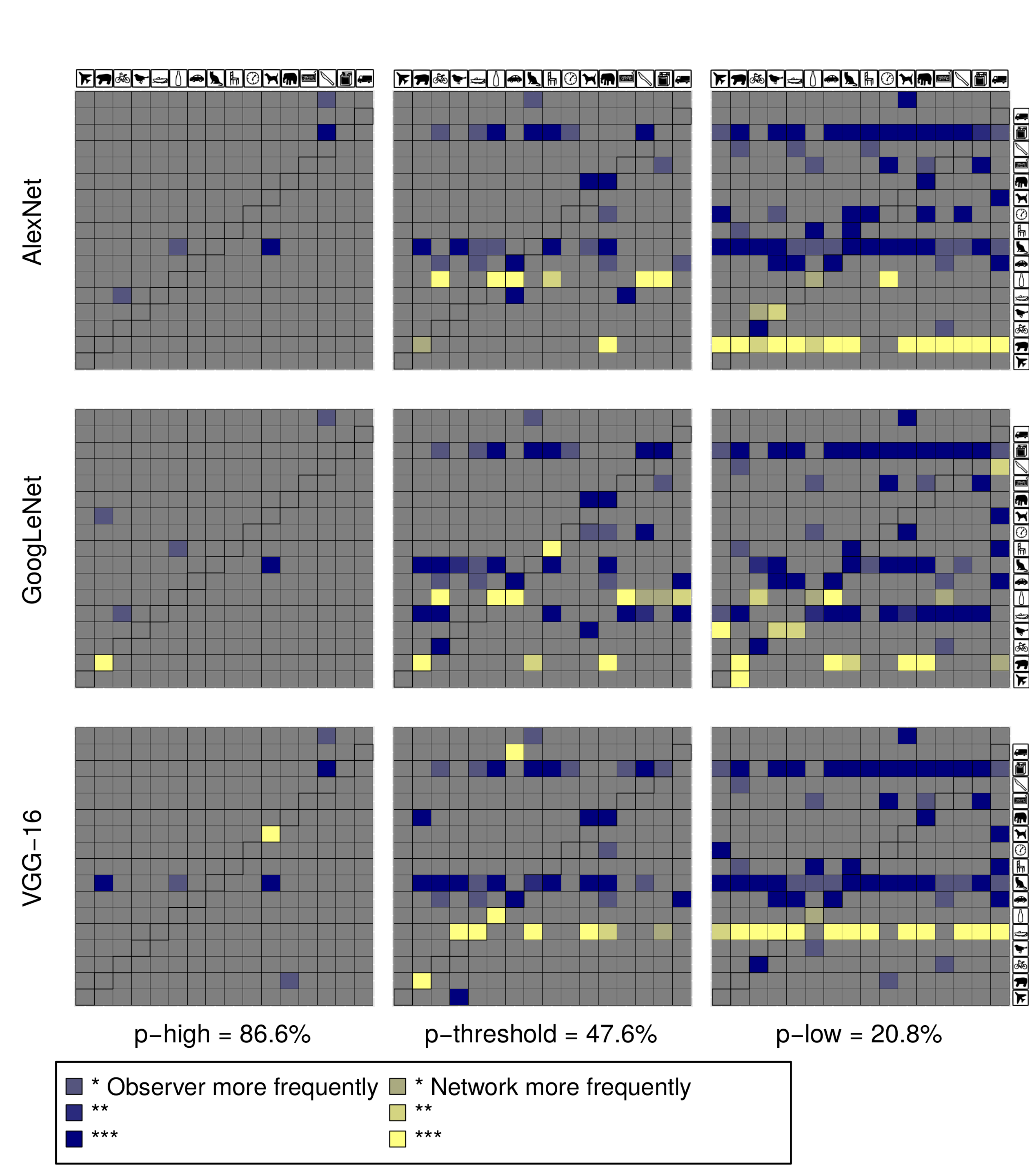}
	\caption{Confusion difference grid for contrast-experiment (as in Figure~\ref{fig:noise_confusion}). Left column: for human observers the condition in which they performed best (100\% contrast, performance p-high = 86.6\%) and for all networks the condition for which they achieved the same accuracy. Right column: data for low performance (20.8\%); middle column: results for medium performance: 47.6\%. The colour here shows whether a difference for a certain cell is significant at $ \alpha = \frac{5\%}{16\cdot 17\cdot 9}(*), \frac{1\%}{16\cdot 17\cdot 9}(**)$ and $\frac{0.1\%}{16\cdot 17\cdot 9}(***) $; these $ \alpha $-levels are Bonferroni corrected for multiple comparisons (16 categories $ \cdot $ 17 possible responses $ \cdot $ 9 confusion difference matrices). Data obtained with JPEG images.}
	\label{fig:contrast_confusion}
\end{figure}

\subsection{Fine-tuning on distortions}
AlexNet, GoogLeNet and VGG-16 have not been designed for or trained on images with reduced contrast, added noise or other distortions. It is therefore natural to ask whether simple architectural modifications or fine-tuning of these networks can improve their robustness. Our preliminary experiments indicate that fine-tuning DNNs on specific test conditions can improve their performance on these conditions substantially, even surpassing human performance on noisy low-contrast images, for example. At the same time, fine-tuning on specific conditions does not seem to generalise well to other conditions (e.g. fine-tuning on uniform noise does not improve performance for salt-and-pepper noise), a finding consistent with results by \citeA{Dodge2017} who examined the impact of fine-tuning on noise and blur. This clearly indicates that it could be difficult to train a single network to reach human performance on all of the conditions tested here. A publication containing a detailed description and analysis of these experiments is in preparation. The question what kind of training would lead to robustness for arbitrary noise models remains open.

\subsection{JPEG vs. PNG}\label{appendix: jpeg_vs_png}
As mentioned in Section~\ref{methods: image_preprocessing}, all experiments were performed using images saved in the JPEG format for compatibility with the image training database ImageNet \cite{Russakovsky2015}, which consists of JPEG images. That is, a certain image was read in, distorted as described earlier and then saved again as a JPEG image using the default settings of the \texttt{skimage.io.imsave} function. Since the lossy compression of JPEG may introduce artefacts, we here examine the difference in DNN results between saving to JPEG and to PNG, which is lossless up to rounding issues. Some results for the contrast-experiment using JPEG images were already reported by \citeA{Wichmann2017}.

\begin{figure}
	\centering
	\subfloat[][Contrast-experiment accuracy for JPEG images]{\includegraphics[width=0.5\textwidth]{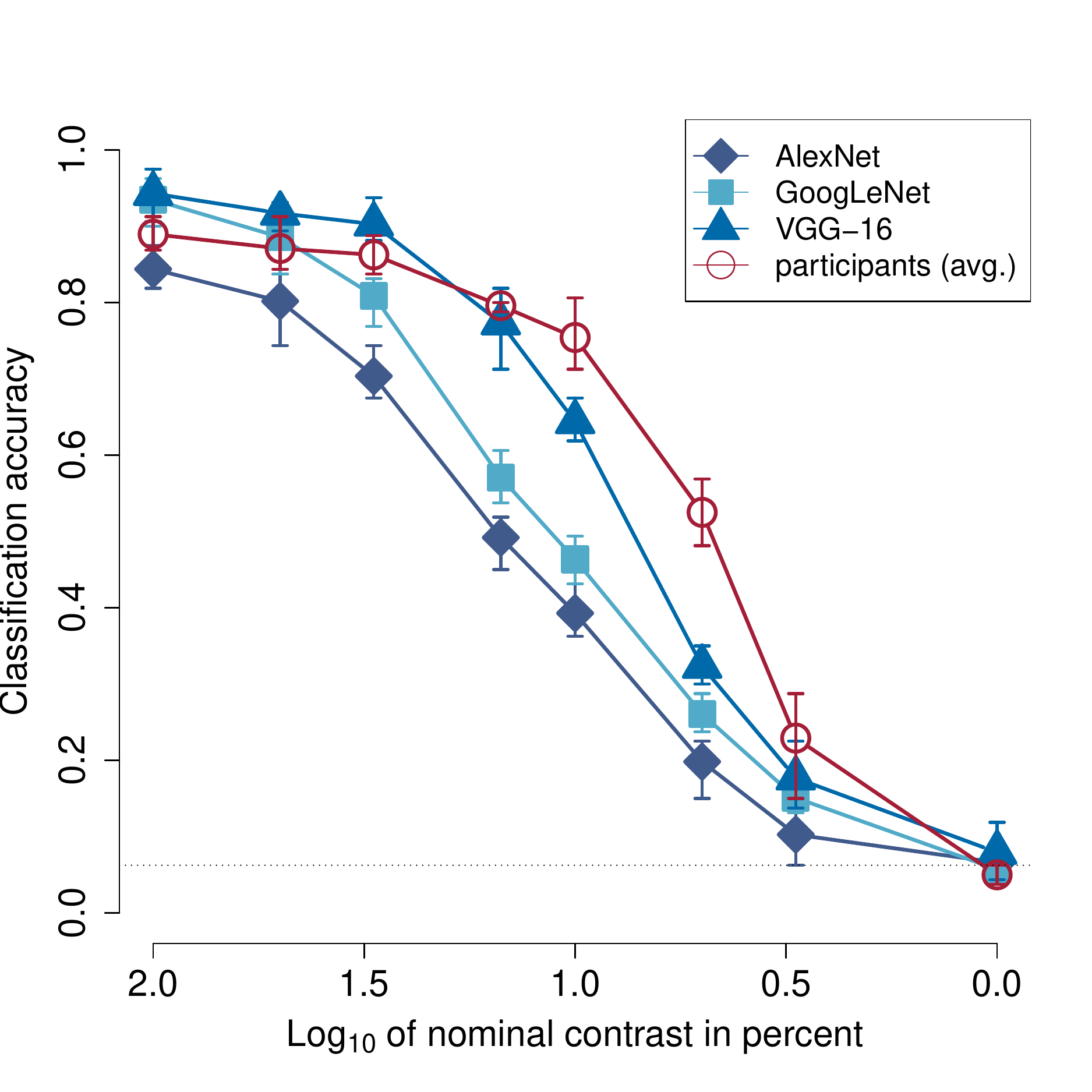}}
	\subfloat[][Contrast-experiment accuracy for PNG images]{\includegraphics[width=0.5\textwidth]{results/contrast/contrast_png_accuracy.pdf}}
	\hfill
	\subfloat[][Contrast-experiment entropy for JPEG images]{\includegraphics[width=0.5\textwidth]{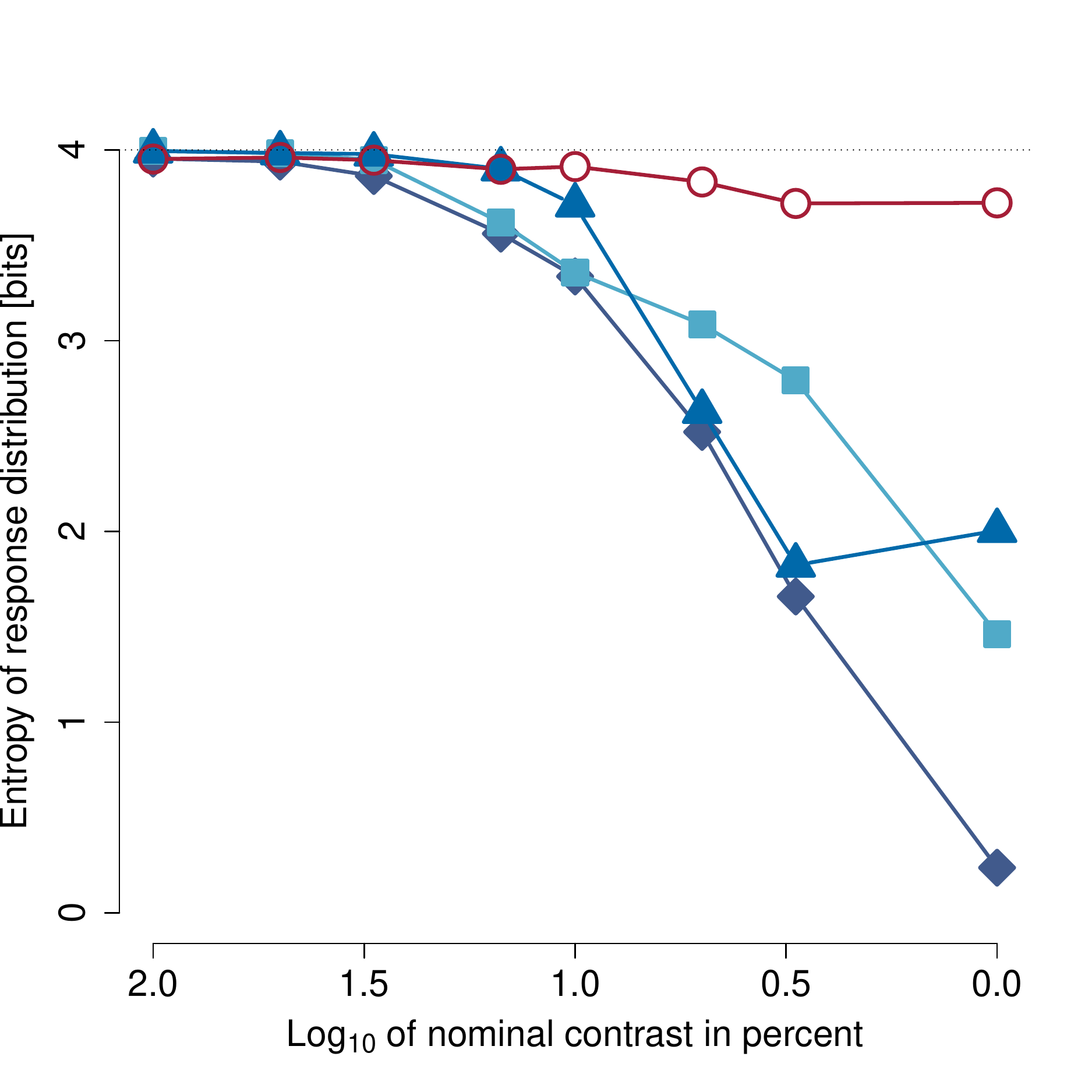}}
	\subfloat[][Contrast-experiment entropy for PNG images]{\includegraphics[width=0.5\textwidth]{results/contrast/contrast_png_entropy.pdf}}
	\caption{Accuracy and response distribution entropy for contrast-experiment, split by JPEG and PNG images. \textbf{(a)(c)} \emph{Results for JPEG images.} N=3, the same three observers as for the PNG experiment. \textbf{(b)(d)} \emph{Results for PNG images.} N=3.}
	\label{fig:jpeg_png_difference_contrast}
\end{figure}

The results of this comparison can be seen in Table~\ref{tab:jpeg_png_difference}. For all experiments but the contrast-experiment, there was hardly any difference between PNG and JPEG images. For the contrast-experiment, however, we found a systematic difference: all networks were better for PNG images. We therefore collected human data for this experiment employing PNG instead of JPEG images (reported in Figure \ref{fig:accuracy_entropy}). In this experiment, three of the original contrast-experiment's observers participated, seeing the same images as in the first experiment\footnote{A time gap of approximately six months between both experiments should minimize memory effects; furthermore, human participants were not shown any feedback (correct / incorrect classification choice) during the experiments.}. The results are compared in Figure~\ref{fig:jpeg_png_difference_contrast}. Both human observers and DNNs were better for PNG images than for JPEG images, especially in the low-contrast regime. Especially VGG-16 benefits strongly from saving images to PNG (on average: 8.82 \% better performance) and achieves better-than-human performance for 1\% and 3\% contrast stimuli. In the main paper, we therefore show the performance of humans and DNNs when the images are saved as in the PNG rather than the JPEG format to disentangle JPEG compression and low contrast.

The cause of this effect could most likely be attributed to JPEG compression artefacts for low-contrast images. Based on our JPEG vs. PNG examination, we draw the following conclusions: First of all, we recommend using a lossless image saving routine for future experiments even though networks may be trained on JPEG images, since performance, as our data indicate, will be either equal or better in both man and machine. Secondly, we showed that our results with JPEG images for the colour-, the noise- and the eidolon-experiment are not influenced by this issue, whereas the contrast-experiment's results are to some degree.

\begin{table}
\begin{centering}
	\caption{Classification performance difference for PNG and JPEG images.}
	\label{tab:jpeg_png_difference}
	\begin{tabular}{lrrrc}
		\toprule
		\multicolumn{1}{c}{Experiment} & \multicolumn{1}{c}{AlexNet} & \multicolumn{1}{c}{GoogLeNet}  & \multicolumn{1}{c}{VGG-16} & \multicolumn{1}{c}{human avg.}\\
		\midrule
		colour-experiment    &  0.03\% (0.03\%)& -0.01\% (0.01\%)&  0.02\% (0.02\%)& - \\
		contrast-experiment &  1.64\% (1.64\%)&  3.25\% (3.27\%)&  8.82\% (8.84\%)& 2.68\% (3.67\%)\\
		noise-experiment    &  0.03\% (0.48\%)&  0.22\% (0.71\%)&  0.45\% (0.71\%)& - \\
		eidolon-experiment  & -0.43\% (1.08\%)&  0.03\% (1.02\%)& -0.34\% (1.09\%)& - \\ \hline
		\bottomrule
	\end{tabular}
\end{centering}
\bigskip\\
\small\textit{Notes}.
Each entry corresponds to the average performance difference for PNG minus JPEG performance for a certain network and experiment. The value in brackets indicates the average absolute difference. A value of 0.03\% for AlexNet in the colour-experiment therefore indicates that AlexNet performance on PNG images was, in absolute terms, 0.03\% higher compared to JPEG images (in this example: 90.58\% vs. 90.61\%). Human data (n=3) was collected for the contrast-experiment only.
\end{table}

\begin{table}
	\centering
	\caption{Colour-experiment: difference between colour and grayscale conditions (paired-samples t-test).}
	\label{tab:colour_difference}
	\begin{tabular}{lccrrc}
		\toprule
		\multicolumn{1}{c}{Network / Observer} & \multicolumn{1}{c}{Difference (\%)} & \multicolumn{1}{c}{95\% CI (\%)} & \multicolumn{1}{c}{\textit{t}} & \multicolumn{1}{c}{\textit{df}} & \multicolumn{1}{c}{\textit{p}} \\
		\midrule
		AlexNet                                & 7.72 & [6.55, 8.90]                         & 12.86               & 4479                   & \textless .001*       \\
		VGG-16                                 & 3.79   & [2.97, 4.62]                       & 8.99                  & 4479                   & \textless .001*       \\
		GoogLeNet                              & 2.90    & [2.04, 3.76]                     & 6.61                  & 4479                   & \textless .001*       \\
		subject-01                             & 0.47       & [-2.91, 3.85]                  & 0.27                  & 639                    & .785                  \\
		subject-02                             & 1.25         &[-2.33, 4.83]                & 0.69                  & 639                    & .493                  \\
		subject-03                             & 3.91           &[0.10, 7.72]              & 2.01                  & 639                    & .045    \\ \hline
		\bottomrule
	\end{tabular}
	\bigskip\\
	\small\textit{Notes}.
	*\textit{p} $ < .008\overline{3}$. Difference stands for colour minus grayscale performance; significant at $\alpha = \frac{5\%}{6} = .008\overline{3}$ after applying Bonferroni correction for multiple comparisons.
\end{table}

\begin{figure}
	\centering
	\subfloat[][Classification accuracy]{\includegraphics[width=0.5\textwidth]{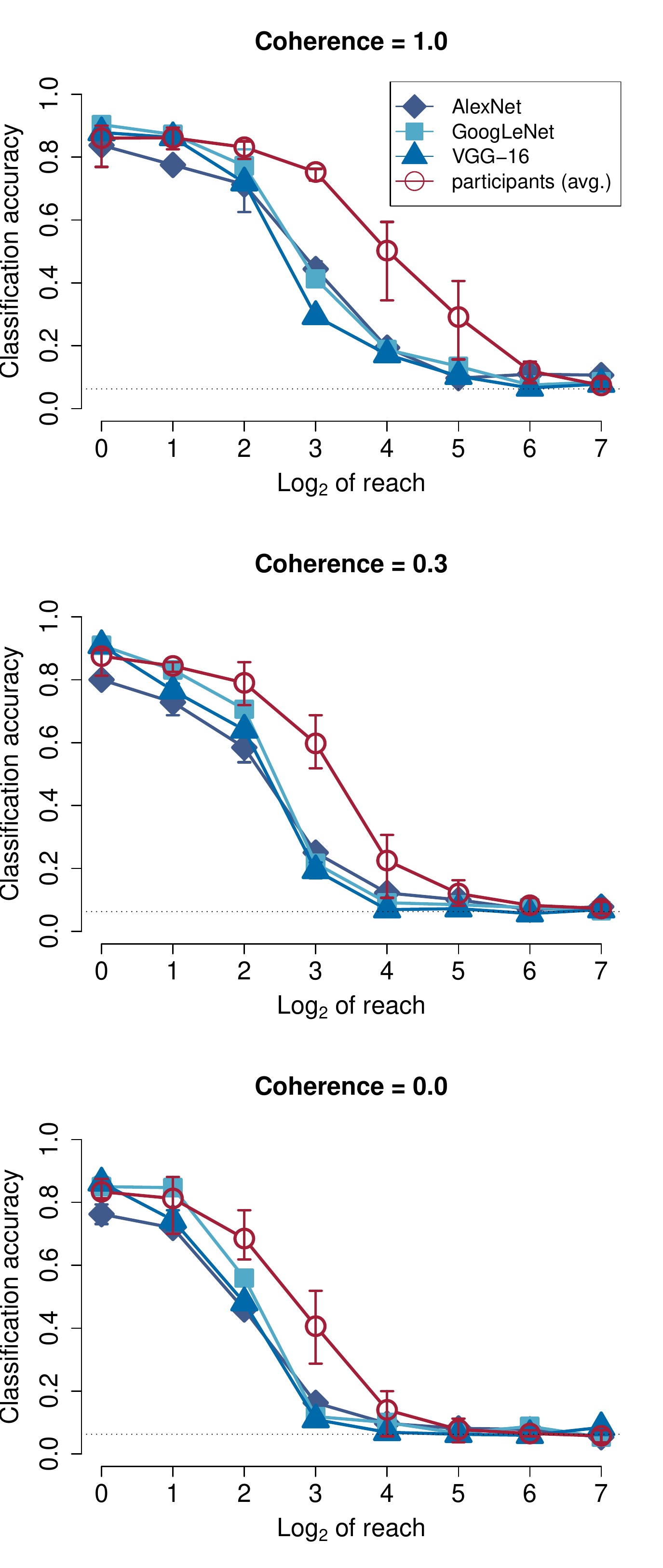}}
	\subfloat[][Response distribution entropy]{\includegraphics[width=0.5\textwidth]{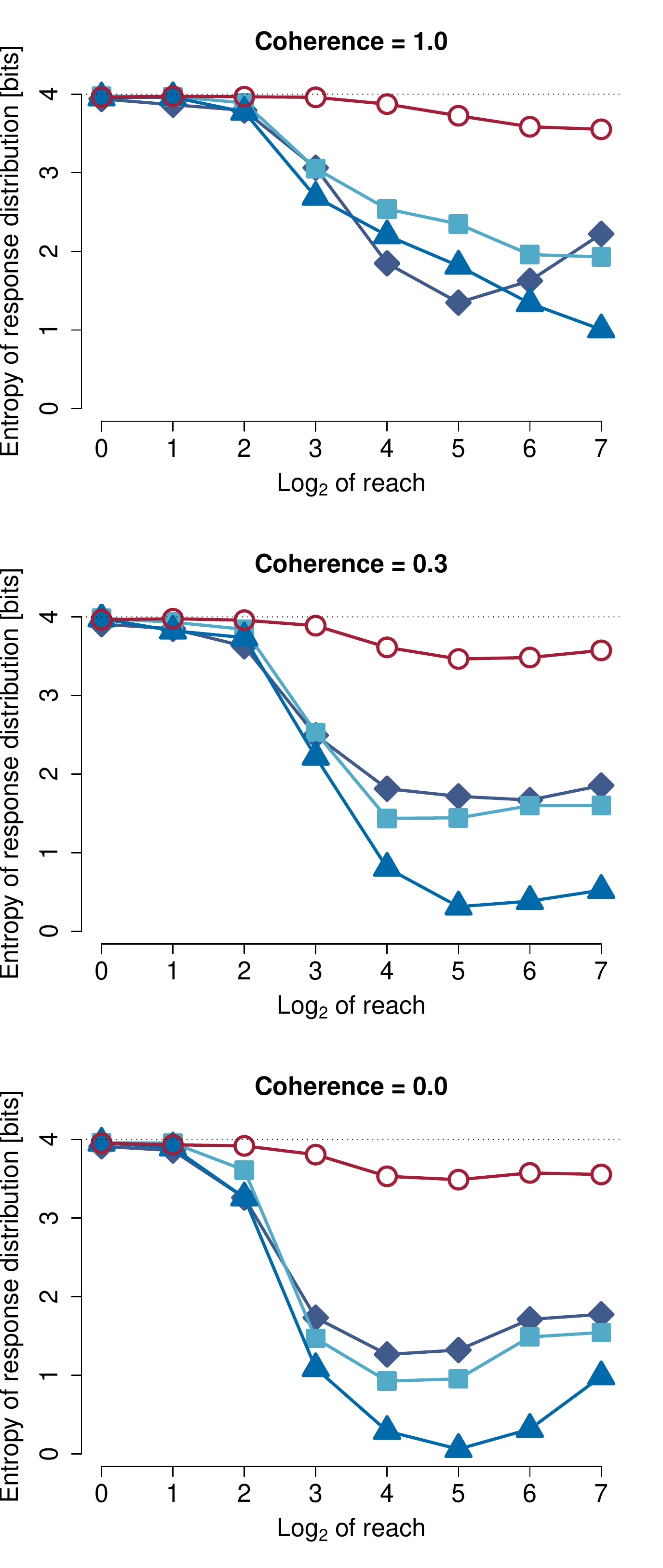}}
	\caption{Complete eidolon-experiment results (n=5). \textbf{(a)} \emph{Accuracy including range}. This range was obtained as for the other experiments. \textbf{(b)} \emph{Response distribution entropy}. Note that the data for coherence = 1.0 are already visualized in Figure \ref{fig:accuracy_entropy} (e) and (f), but are here shown again for better comparison to the results obtained for different coherence values.}
	\label{fig:eidolon_performance}
\end{figure}

\begin{figure}
	\centering
	\subfloat[][Contrast-experiment, JPEG images]{\includegraphics[width=0.50\textwidth]{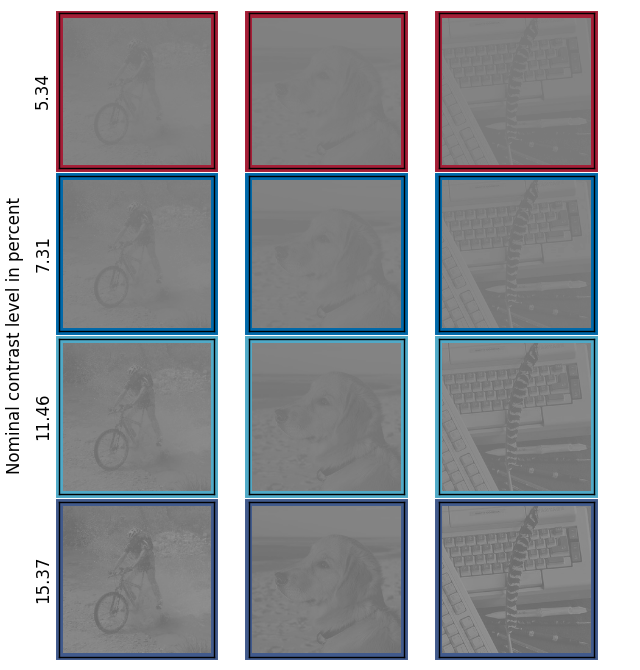}}
	\subfloat[][Contrast-experiment, PNG images]{\includegraphics[width=0.50\textwidth]{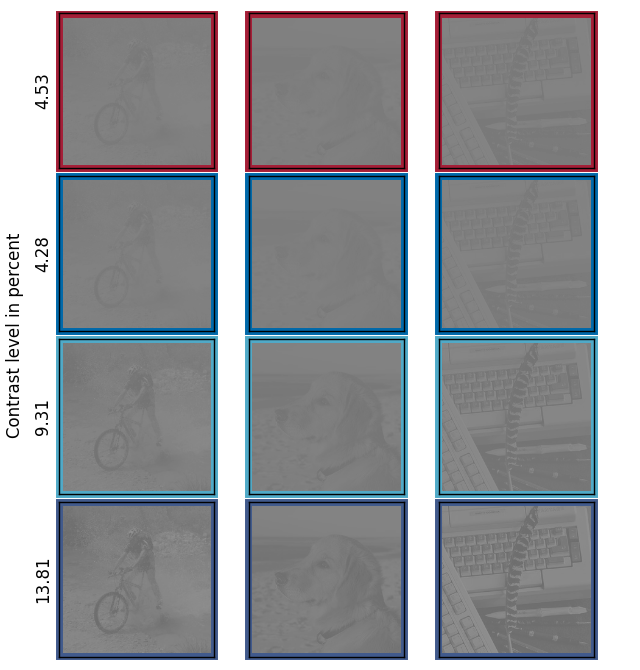}}
	\hfill
	\subfloat[][Eidolon-experiment, coherence = 0.3]{\includegraphics[width=0.50\textwidth]{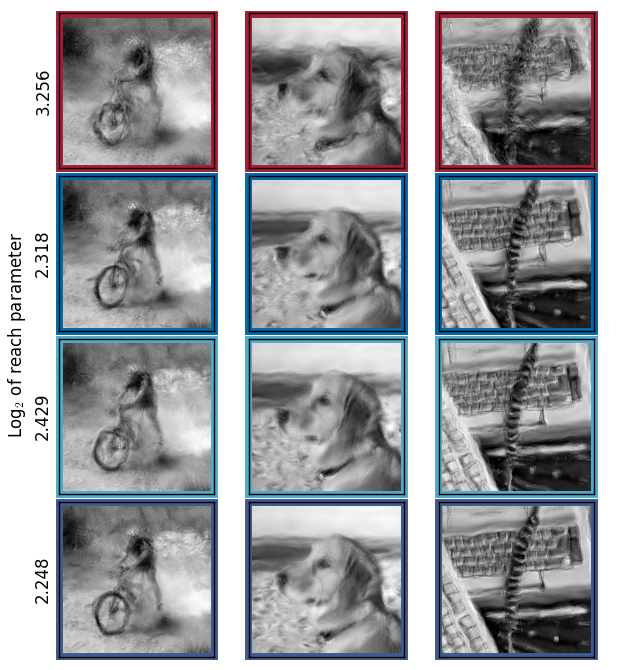}}
	\subfloat[][Eidolon-experiment, coherence = 0.0]{\includegraphics[width=0.50\textwidth]{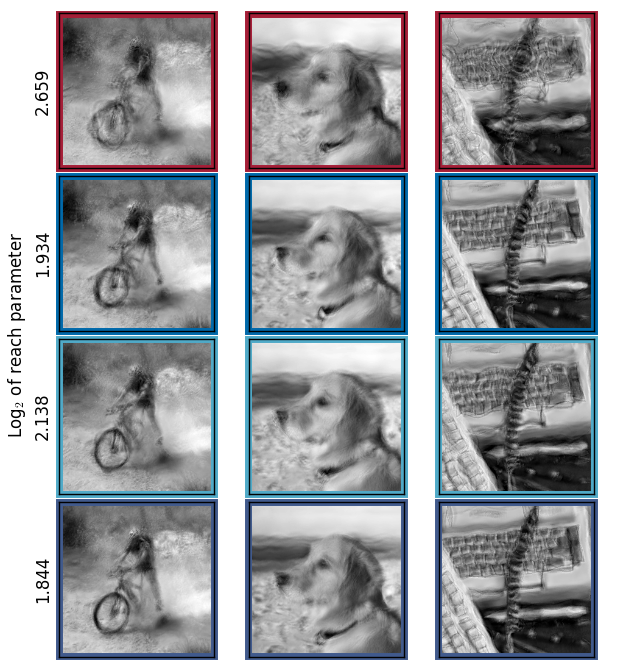}}
	\caption{Estimated stimuli corresponding to $ 50\% $ accuracy for contrast-experiments and eidolon-experiment (coherence parameter = 0.3 and 0.0). Top row: stimuli corresponding to threshold for the \textcolor{human.100}{average human observer}. Bottom three rows: stimuli corresponding to the same accuracy for \textcolor{vgg.100}{VGG-16} (second row), \textcolor{googlenet.100}{GoogLeNet} (third row) and \textcolor{alexnet.100}{AlexNet} (bottom row). The corresponding stimulus levels were calculated by assuming a linear relationship between the two closest data points measured in the experiments.}
	\label{fig:threshold_contrast_eidolon}
\end{figure}

\end{document}